\documentclass[final]{cvpr}

\usepackage{times}
\usepackage{epsfig}
\usepackage{graphicx}
\usepackage{amsmath}
\usepackage{amssymb}

\usepackage[utf8]{inputenc} 
\usepackage[T1]{fontenc}    
\usepackage{url}            
\usepackage{booktabs}       
\usepackage{amsfonts}       
\usepackage{nicefrac}       
\usepackage{microtype}      

\usepackage{subfig}
\usepackage{tabularx}
\usepackage{graphicx}

\usepackage{xcolor}

\usepackage{algorithm}      
\usepackage[noend]{algpseudocode} 

\usepackage[pagebackref=true,breaklinks=true,colorlinks,bookmarks=false]{hyperref}

\def\cvprPaperID{6288} 
\def\confYear{CVPR 2021}

\begin{document}


\title{On the Difficulty of Membership Inference Attacks}


\author{Shahbaz Rezaei \space\space\space\space\space\space\space Xin Liu \space\space\space \\
University of California\\
Davis, CA, USA \\
{\tt\small \{srezaei,xinliu\}@ucdavis.edu} \\

}


\maketitle

\begin{abstract}
  Recent studies propose membership inference (MI) attacks on deep models, where the goal is to infer if a sample has been used in the training process. Despite their apparent success, these studies only report accuracy, precision, and recall of the positive class (member class). Hence, the performance of these attacks have not been clearly reported on negative class (non-member class). In this paper, we show that the way the MI attack performance has been reported is often misleading because they suffer from high false positive rate or false alarm rate (FAR) that has not been reported. FAR shows how often the attack model mislabel non-training samples (non-member) as training (member) ones. The high FAR makes MI attacks fundamentally impractical, which is particularly more significant for tasks such as membership inference where the majority of samples in reality belong to the negative (non-training) class. Moreover, we show that the current MI attack models can only identify the membership of misclassified samples with mediocre accuracy at best, which only constitute a very small portion of training samples. 
  
  We analyze several new features that have not been comprehensively explored for membership inference before, including distance to the decision boundary and gradient norms, and conclude that deep models' responses are mostly similar among train and non-train samples. We conduct several experiments on image classification tasks, including MNIST, CIFAR-10, CIFAR-100, and ImageNet, using various model architecture, including LeNet, AlexNet, ResNet, etc. We show that the current state-of-the-art MI attacks cannot achieve high accuracy and low FAR at the same time, even when the attacker is given several advantages. 
  The source code is available at \href{https://github.com/shrezaei/MI-Attack}{\color{blue} {https://github.com/shrezaei/MI-Attack}}.
\end{abstract}

\section{Introduction}

There is an extensive recent literature on membership inference (MI) attacks on deep learning models that achieve high MI attack accuracy \cite{shokri2017membership, salem2018ml, liu2019socinf, song2019privacy, long2017towards, truex2019demystifying, long2018understanding, yeom2018privacy}. These MI attack models often use confidence values of the target model to infer the membership of an input sample. High MI attack accuracy is often justified by claiming that deep learning models are more confident towards the training (member) samples than the samples they have not seen during training\footnote{In this paper, we use training samples, member samples, and positive samples interchangeably. Likewise, we also use non-training, test, non-member, and negative samples interchangeably.}. Consequently, MI attack accuracy is reported to be highly correlated to model's overfitting or generalization gap \cite{salem2018ml, shokri2017membership, song2019privacy} because an overfitted model should perhaps behave even more confident towards training samples. 

\begin{table}
\footnotesize
  \caption{Complete evaluation of CIFAR-100 with three different target models. Almost all papers report the third section that includes accuracy, precision, recall, and F1 score. The second section, including train and test accuracy of the target victim model, is missing in many papers in literature despite its usefulness in evaluating the generalization gap (and degree of overfitting). The last section that includes balanced accuracy and FAR has never been reported, but it is of paramount importance for understanding the performance of attack models in practice.}
  \label{cifar-100-reproduced-tbl}
  \centering
  \begin{tabular}{llllll}
    \toprule
    \midrule
    Dataset  & Cifar-100  & Cifar-100 & Cifar-100 \\
    Model  & AlexNet  & ResNet & DenseNet \\
    \midrule
    Target Model Train Acc. & 92.48\% & 95.80\% & 99.98\% \\
    Target Model Test Acc. & 43.87\% & 74.14\% & 82.83\%  \\
    \midrule
    Attack Acc. & 82.62\% & 79.13\% & 87.74\%  \\
    Attack Precision & 91.90\% & 87.3\% & 86.97\% \\
    Attack Recall & 86.92\% & 87.85\% & 98.29\% \\
    Attack F1 & 89.23\% & 87.45\% & 92.26\% \\
    \midrule
    Attack Bal. Acc. & 74.02\% & 61.70\% & 66.65\% \\
    \textbf{Attack FAR}  & 38.89\% & 64.45\% & 65.00\% \\
    \bottomrule
  \end{tabular}
\end{table}

In this paper, we show that the way the previous papers report the attack performance do not reveal how exactly these attacks perform in practice and can be misleading. First, many papers do not provide the train and test accuracy of the target victim models. Hence, it is not clear whether the target model is well-trained. We only find a handful of papers that report such metrics \cite{shokri2017membership, long2017towards, nasr2019comprehensive, jayaraman2019evaluating, jia2019memguard}. Even in these cases, one can clearly spot impractical target models where generalization gap is sometimes larger than 35\% \cite{shokri2017membership, long2017towards}, 50\% \cite{nasr2019comprehensive}, or 80\% \cite{jayaraman2019evaluating}. Clearly, such extremely overfitted models have no practical use and the results on such models should not be generalized to well-trained models. Second, all papers we have examined limit their reporting to accuracy, precision, and recall.
Such a reporting does not reveal the performance of attack models on negative samples (non-member), especially how many negative samples are misclassified as positive (false positive). 
For binary classification tasks, this is of crucial importance, especially when the negative class can significantly outnumber the positive class (e.g., all possible images vs.~the limited number of images used to train an image classification model). A good practice, which is also common in other fields such as intrusion detection systems \cite{wu2018novel}, is to report false positive rate (FPR) or false alarm rate (FAR) alongside the other metrics. A good attack model should have a low FAR. 


To evaluate the feasibility of MI attack, we tried to reproduce the results in \cite{nasr2019comprehensive} for CIFAR-100 dataset, presented in Table \ref{cifar-100-reproduced-tbl}\footnote{In literature, we have only found two papers with public source codes \cite{song2019privacy, salem2018ml}. We run their implementation on CIFAR-10 and observed the same problem. They both suffer from high FAR, which has not been reported before.}. Although the generalization gap of AlexNet is high ($\sim 48\%$), we keep the results for the sake of comparison. As it is shown, commonly used metrics, including accuracy, precision, and recall, do not reveal how attack models really perform on non-member samples. The high FAR of these attacks make them unreliable. Interestingly, even the attack on an extremely overfitted model such as AlexNet still suffers from high FAR.

In this paper, we first elaborate on why previous reporting practices are misleading in membership inference research. Second, we provide a comprehensive evaluation of membership inference attacks on deep learning models. We give as much advantage as possible to an attacker and we show that a reliable MI attack with high accuracy and low FAR is hard to achieve. We show that the reason MI attacks often fail is not because attack models are trained poorly. The reason is that the statistical properties the features used in MI attacks are not clearly different and distinguishable for training and non-training sample. 

To provide an insight on why membership inference of some samples are possible, we separate datasets into two parts: correctly classified samples (by the target victim model) and misclassified samples. In general, we find that membership inference of correctly classified samples, independent of what dataset or model is used, is a more difficult task than the membership inference of misclassified samples. This is because deep learning models often behave similarly on train and non-train samples when they are correctly classified. This observation sheds light on the difficulty of membership inference on deep models.

Our contributions are summarized as follows:
\begin{itemize}

\item We show that attack accuracy, precision, and recall are not enough to show the performance of MI attacks, particularly on negative (non-member) samples. Instead, we should also report FAR (or other substitutes explained in Sec.\ref{sec-reporting}) and train/test accuracy of target models to better demonstrate the performance of MI attacks. Moreover, we study the performance of correctly classified samples and misclassified samples separately. We show that membership inference of correctly classified samples, to which the majority of training samples belong, is a very difficult task.

\item We perform MI attack on various image datasets (including MNIST, CIFAR-10, CIFAR-100, and ImageNet), and models (LeNet, AlexNet, ResNet, DenseNet, InceptionV3, Xception, etc), some of which are studied for the first time in the MI context. We conduct experiments such that they give a lot more advantages to the attacker than in any previous work. Even in this case, we show that a meaningful membership inference attack with high accuracy and low FAR is often not achievable.

\item In addition to confidence values of the target (victim) model, we extensively analyze and use other information available from the target model, including values from intermediate layers, the gradient w.r.t input, gradient w.r.t to model weights, and distance to the decision boundary. In some cases, these types of information slightly leak more membership status than confidence values, but they are still not sufficient for a reliable MI attack in practice. Surprisingly, all evidence suggests that deep models often behave similarly on train and non-train samples across all these metrics. The only considerable difference appears between correctly classified samples and misclassified samples, not between the train and non-train samples.

\end{itemize}

\begin{table*}
\scriptsize
  \caption{Performance evaluation of an MI attack model when balancedness of the evaluation set is changed.}
  \label{tbl-cifar100-resnet}
  \centering
  \begin{tabular}{llllllll}
    \toprule
    Balancedness & Attack Model & \multicolumn{2}{c}{Positive (member) Class} & \multicolumn{2}{c}{Negative (non-member) Class}  & Balanced Accuracy & FAR \\
    \cmidrule(r){3-4}
    \cmidrule(r){5-6}
    -     & - & Precision & Recall & Precision & Recall & - & - \\
    \midrule
    5:1 & MI Attack & 87.30\% & 87.85\% & 38.18\% & 35.55\% & 61.70\% & 64.45\% \\
    5:1   & ZeroR & 83.33\% & 100.0\% & 0.0\% & 0.0\% & 50.0\% & 100.0\% \\
    \midrule
    1:1 & MI Attack & 57.68\% & 87.42\% & 74.49\% & 35.42\% & 61.22\% & 64.82\% \\
    1:1   & ZeroR & 50.0\% & 100.0\% & 0.0\% & 0.0\% & 50.0\% & 100.0\% \\
    \midrule
    1:5 & MI Attack & 21.41\% & 87.82\% & 93.57\% & 35.73\% & 61.28\% & 64.42\% \\
    1:5   & ZeroR & 16.66\% & 100.0\% & 0.0\% & 0.0\% & 50.0\% & 100.0\% \\
    \bottomrule
  \end{tabular}
\end{table*}

In summary, our experiments, including the reproduction of results in the literature, suggest membership inference of correctly classified samples, to which the majority of training samples belong, is a difficult task. Clearly more research is needed and we are hesitant to generalize our results to all scenarios, some of which are as follows:

\begin{itemize}
\item We mainly focus on vision tasks with high dimensional input. Membership inference for other tasks with low dimensional input may culminate in a different result.

\item We do not extend our conclusion to generative models. High capacity generative models can often memorize training samples, which can be retrieved at inference time, as shown in \cite{carlini2019secret}. However, there is no trivial method to retrieve memorized samples from a discriminative model even if it memorizes training samples.

\item We do not extend the conclusion to any attack that can be launched during the training phase, such as data poisoning, model/training manipulation \cite{song2017machine}, etc. We only study membership inference on naturally trained models and natural datasets. Deep models may behave very differently if any unnatural manipulation appears during the training phase.

\item In each dataset, there often exists outliers. The MI attack maybe more successful on these samples \cite{long2018understanding}, whether they are classified correctly or not.

\end{itemize}

\section{Related Work}

The first membership inference attack on deep models is proposed by Shokri et al. in \cite{shokri2017membership}. The key idea is to build a machine learning attack model that takes the target model's output (confidence values) to infer the membership of the target model's input. To train the attack models, membership dataset containing $(x_{conf}, y_{mem})$ pairs is needed where $x_{conf}$ represents the confidence values obtained by the target model for each sample and $y_{mem}$ is a binary variable indicating whether the sample is used in target model's training or not. To build the membership dataset, a set of shadow models are trained for which the training and non-training samples are known. The attack is possible under two assumptions \cite{salem2018ml}: (a) the shadow models share the same structure as the target model, and (b) the training dataset used to train the shadow models share the same distribution as the one used to train the target model. To mitigate these limitations, Salem et al. \cite{salem2018ml} relax the second assumption by showing the attack is possible using different datasets and the first assumption by proposing a threshold-based attack that does not require a training procedure. To further relax these assumptions, several studies \cite{liu2019socinf, song2019privacy, long2017towards, truex2019demystifying, long2018understanding, yeom2018privacy} introduce better dataset generating procedures for shadow models, and extend the experiments to various scenarios and datasets. However, all studies share the same idea of using target model's output for membership inference. In this paper, we show that such attacks that rely on confidence values either suffer from high FAR or low accuracy. Moreover, we show that the output of deep learning models are often similar for correctly labeled samples, whether they were used in training or not.

\begin{figure*}
\centering
\centering
\begin{tabular}{cc}
\subfloat[Correctly labeled]{\includegraphics[width=0.3\linewidth]{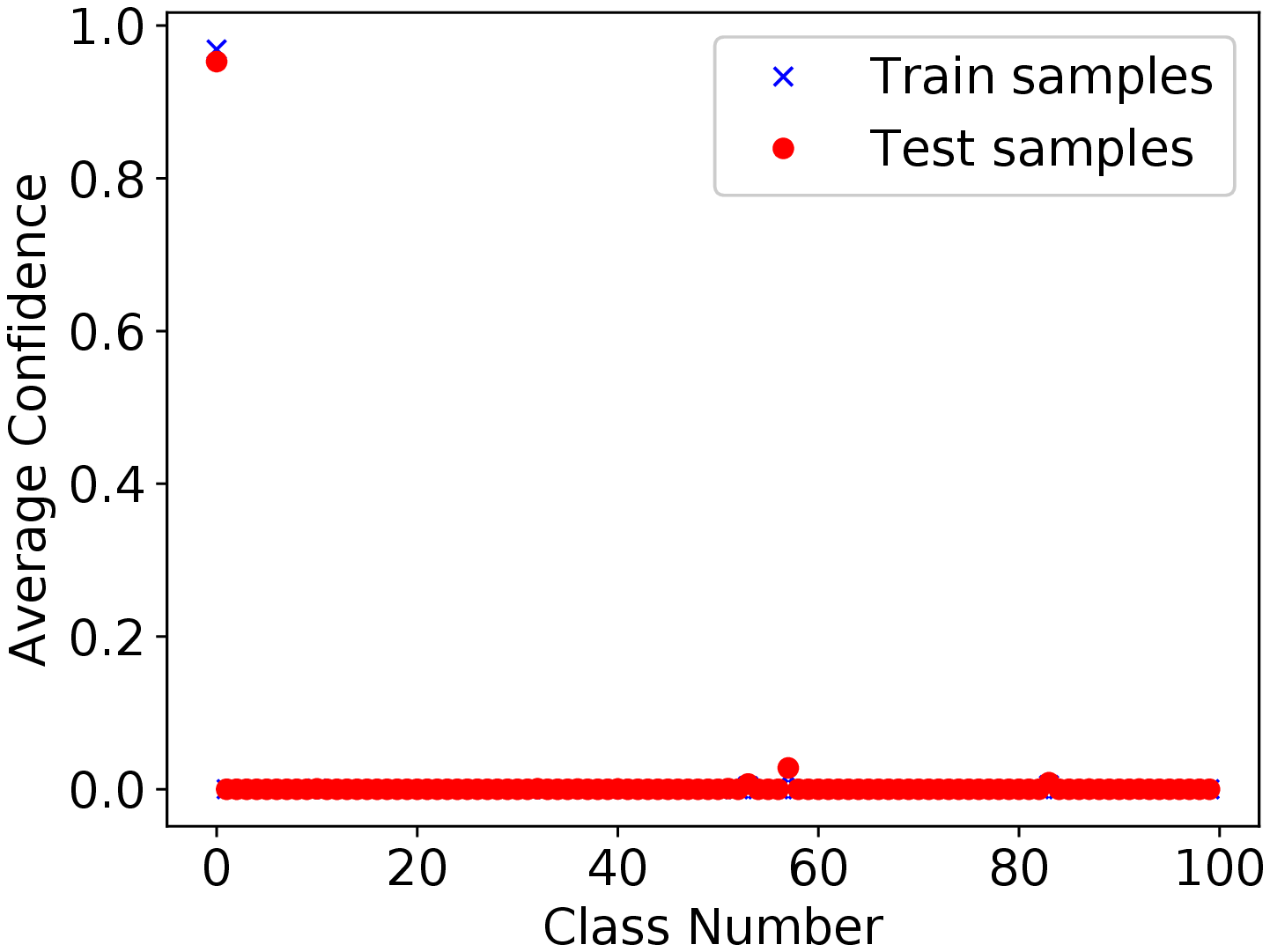}} 
& \subfloat[Misclassified]{\includegraphics[width=0.3\linewidth]{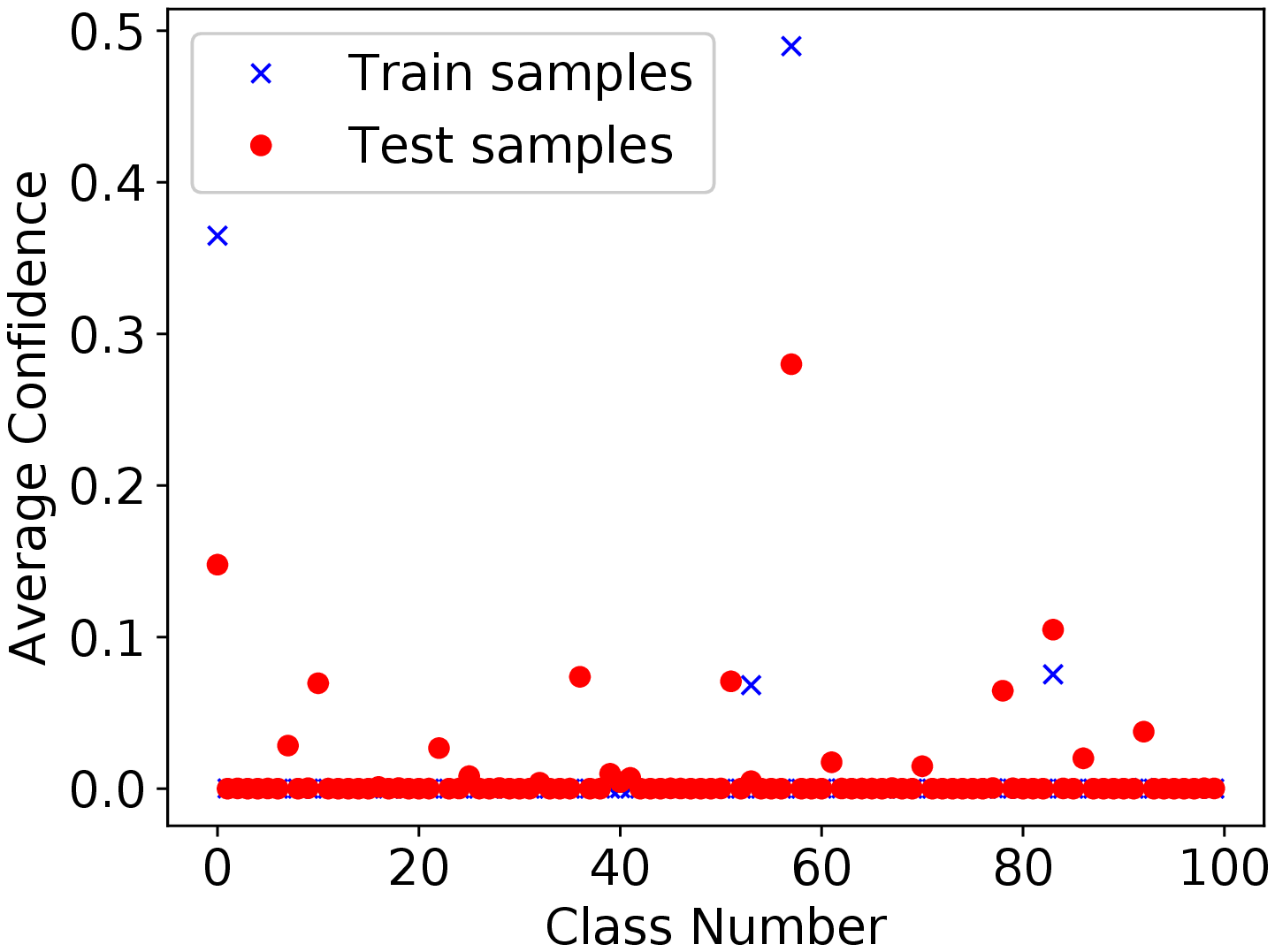}}\\
\end{tabular}
\caption{Distribution of average confidence values for CIFAR 100 dataset (ResNet) (class 0) }\label{conf-dist-resnet}
\end{figure*}

\section{Better MI Attack Reporting}
\label{sec-reporting}
As discussed in Section I, the common approach of reporting accuracy, precision, and recall of MI attacks does not truly reveal the performance of them on negative (non-member) samples. By providing false alarm rate (FAR) in Table \ref{cifar-100-reproduced-tbl}, we show that these attacks suffer from high false positive ratio. The reason why high false positive ratio does not significantly affect the reported precision ($\frac{TP}{TP+FP}$) is due to the MI dataset imbalancedness. In a typical machine learning training, majority of samples are used for training and, consequently, the holdout (test) set is considerably smaller. For instance, the training:test (or member:non-member) ratio of CIFAR dataset is 5:1 and, consequently, the MI dataset to train/evaluate the MI attack model has the same imbalancedness ratio. As a result, the total number of FPs is small despite the high false positive ratio. This is clearly illustrated in Table \ref{tbl-cifar100-resnet} where the member:non-member ratio varies from 5:1, 1:1, to 1:5 and the precision on the positive class dropped from 87\% to 21\% for the same MI attack model. Note the MI attack model is the same for all experiments which was trained on a balanced dataset. We only change the balancedness of the evaluation set in Table \ref{tbl-cifar100-resnet}.

We emphasize that reporting performance only on positive samples can be misleading. 
In Table \ref{tbl-cifar100-resnet}, we report the precision and recall for both positive and negative classes. When the 5:1 ratio of balancedness is kept, the MI attack shows high precision and recall on positive samples, but low precision and recall on negative samples (which in general has not been reported). To stress this message, we also report ZeroR as a baseline. ZeroR is a classifier that always predicts the positive class (member). As shown in  Table \ref{tbl-cifar100-resnet}, ZeroR performance is high and close to the MI attack when {\bf only} the precision and recall of positive samples are considered. This clearly shows that one should also report performance on negative samples as well.

The training:test ratio can significantly change the performance metrics of a same model, as shown in Table \ref{tbl-cifar100-resnet}. So what is the right way to treat this ratio? Previous MI papers often keep the radio of 5:1, or in some cases 1:1 \cite{nasr2019comprehensive}. However, in practice, the ratio can reverse drastically: for example, a random sample, chosen from the distribution of all natural images, has a significantly higher probability of being a non-member sample than a member sample. 
However, there is no practical way to estimate the true value of this ratio. Furthermore, even if the ratio is known, due to the limited number of samples available for evaluation, it might not be  practical to obtain more non-member samples to achieve the true ratio. 
As a result, in this paper, we keep the ratio as 5:1, as in most existing papers in the evaluation set. Instead, we report performance metrics that are less sensitive to the balancedness ratio, i.e, balanced accuracy and FAR\footnote{ A more elaborated argument for why FAR is important in tasks where one class hugely outnumber the other class is the base-rate fallcay \cite{axelsson2000base}.}.

There is another way to show the ineffectiveness of current MI attacks using a rather simple baseline. In \cite{leino2020stolen}, such a baseline is introduced,  called naive attack. The idea is to predict a sample as member if it is correctly labeled by the target model, and to predict it as non-member if misclassified. Clearly, the FAR of naive attack is high because it classifies all correctly classified samples as members. However, this impractical attack can also achieve high accuracy on positive samples. Since this naive attack is obviously ineffective in practice, one can compare the accuracy gap between the naive attack and a MI attack to conclude the effectiveness of an MI attack. This approach has been used in \cite{leino2020stolen}. We report the accuracy of this naive attack for completeness in Table \ref{target-model-accuracy-tbl}, but we rely on accuracy/FAR pair to evaluate an attack's success.

Despite their low performance, MI attacks still outperform the random guess. To shed light on why MI attacks are effective on some samples, we report the behavior of target models on correctly classified and misclassified samples, separately. We show that deep learning models often behave similarly when they correctly label a sample, whether it is a training (member) or test (non-member) sample. In comparison, deep learning models demonstrate slightly different behavior on misclassified samples, which can be exploited by MI attack models. Hence, we believe that separating correctly classified and misclassified samples, and reporting the MI attack on them separately provides a better insight on how MI attacks work.

In summary, the way MI attack has been reported in literature does not provide a complete picture of their performance. Relying only on precision and recall of the positive class can present a delusion of successful attack, particularly when the imbalancedness issue is ignored. Instead, we should report performance on both positive and negative samples, i.e., adding  FAR, or precision and recall on negative samples.
Furthermore, simple baselines, such as ZeroR and naive attack, can be used for comparison. Last, separating correctly labeled and incorrectly labeled samples provide a better insight on how these MI attacks work. Therefore, in this paper, we report balanced accuracy and FAR, and we also report the performance on correctly classified and misclassified samples, separately, when possible.

\section{Methodology}
\label{sec-methodology}
\paragraph{Threat model and assumptions}
In this paper, we give an attacker as much advantage as possible to show that even in such cases membership inference cannot significantly outperform the baseline. We assume a white-box access to the model and unlimited number of queries. Moreover, we give the membership status of up to $80\%$ of training samples and test samples to the attackers and we only ask the attack model to predict the membership inference of the remaining samples. Hence, the attack performance we report in this paper is as good as or better than any proposed attack based on shadow models \cite{shokri2017membership, long2017towards} or transferred or synthesized data \cite{liu2019socinf, salem2018ml} \footnote{ Note that the use of these methods are beneficial when the dataset is small for training a MI attack because one can potentially multiply the MI training dataset by obtaining multiple shadow models. However, in our study, such methods are not necessary since MNIST, CIFAR, and ImageNet datasets have abundant samples.}. In addition to confidence values of the target model, which have been used extensively for membership inference attack in the past, we also study the output of intermediate layers, distance to the decision boundary and a set of gradient norms to better understand if deep models behave differently on training and test samples.

\paragraph{Confidence values}
Confidence values, or the output of Softmax layer, have been widely used for membership inference \cite{shokri2017membership,salem2018ml, liu2019socinf, song2019privacy, long2017towards, truex2019demystifying, long2018understanding, yeom2018privacy}. Figure \ref{conf-dist-resnet} shows the distribution of average confidence for correctly classified samples and misclassified samples of a ResNet model trained on CIFAR-100. As it is shown, misclassified samples often show different distribution for training samples and non-training samples. However, correctly classified samples often saturate the true class confidence value and zero out other confidence values. We show in Section \ref{eval} that membership inference attack models are often fail to considerably outperform a coin toss for correctly labeled samples.

\paragraph{Output of intermediate layers} In deep models, first layers often extract general and simple features that are not specific to training samples. As suggested in \cite{nasr2019comprehensive}, the last layer and layers close to the last one contain more sample-specific information. Hence, we also examine the output of the fully connected layers before the Softmax for membership inference attacks.

\begin{algorithm}
\footnotesize
\caption{FGM-based algorithm to find distance to the decision boundary}
\label{algo-distance}
\begin{algorithmic}[1]
 \Require{$S$ (maximum number of steps), $\textbf{x}$ (input sample) and $y$ (the sample ground-truth), $f_{cls}$ (an interface to the target model which returns predicted class), $f_{conf}$ (an interface to the target model which returns the confidence value of the predicted class), $L$ (target model loss function), $\theta$ (confidence threshold indicating when the algorithm stops optimizing):}
\Procedure{Distance\_To\_Boundary}{$\textbf{x}$, $f$}
\State $\textbf{x}_0 = \textbf{x}$
 \For{t from $0$ to $S$}
  \State $\textbf{x}_{t+1} = \textbf{x}_{t} + \varepsilon \frac{\nabla_x L(x_t,y)}{|| \nabla_x L(x_t,y) ||_2}$
    \If{$f_{cls}(\textbf{x}_{t+1}) \neq f_{cls}(\textbf{x}_{t})$}
    \While{$|f_{conf}(\textbf{x}_{t+1}) - f_{conf}(\textbf{x}_{t})| > \theta $ } 
    \State $\textbf{x}_{m} = \frac{\textbf{x}_{t+1} + \textbf{x}_{t}}{2}$
        \If{$f_{cls}(\textbf{x}_{m}) = f_{cls}(\textbf{x}_{t})$}
        \State $ \textbf{x}_{t} = \textbf{x}_{m} $
        \ElsIf{$f_{cls}(\textbf{x}_{m}) = f_{cls}(\textbf{x}_{t+1})$}
        \State $ \textbf{x}_{t+1} = \textbf{x}_{m} $
        \Else
        \State{ \Return \textit{Error!} }
        \EndIf
    \EndWhile
   \Return $ || \textbf{x}_0 - \textbf{x}_t ||_2$
   \EndIf
 \EndFor
 \Return \textit{Optimization failed!}
 \EndProcedure
 \end{algorithmic}
\end{algorithm}

\paragraph{Distance to the decision boundary}
Some research focuses on understanding decision boundary of deep models \cite{mickisch2020understanding, karimi2019characterizing} or geometry and space of deep models \cite{fawzi2018empirical, moosavi2019robustness} to often understand the nature of adversarial examples or to improve robustness. 
In this paper, we investigate whether the distance to boundary is a distinguishable feature for membership inference. To find the distance to the decision boundary, we use FGM \cite{dong2018boosting} optimization procedure to craft an image on the other side of the decision boundary (Algorithm \ref{algo-distance}). Then, we perform a binary search to find an instance for which the model's confidence for two classes are almost equal, that is, the difference between two confidences is smaller than a small threshold, similar to \cite{karimi2019characterizing}. Finally, we obtain the $L_2$ distance between the original sample and the crafted samples as a measure of distance to the boundary.

\paragraph{Gradient norm}
It has been shown that the gradient of loss with respect to model parameters, $\frac{\partial L}{\partial w}$, is often smaller for training samples than non-training samples \cite{nasr2019comprehensive} and it can be used for membership inference attack in federated learning scenario. In this paper, we study the gradient of loss with respect to model parameters, $\frac{\partial L}{\partial w}$, and also the gradient of loss with respect to model input, $\frac{\partial L}{\partial x}$, in a non-federated learning setting. The large value for the former indicates that major re-tuning of model parameters is needed for that sample, and hence, it can be an indication of a non-member sample. The large value of the latter indicates that there are input samples with more confident output in the vicinity of that sample, and hence, it can be an indication of a non-training sample. Both $\frac{\partial L}{\partial w}$ and $\frac{\partial L}{\partial x}$ are extremely high dimensional. Thus, we adopt the seven norms used in \cite{oberdiek2018classification}, originally used for analysis of deep model's uncertainty, namely $L_1$, $L_2$, absolute minimum, $L_\infty$, mean, Skewness, and Kurtosis.

\begin{table}
\scriptsize
  \caption{Accuracy of various datasets, target models, and MI attack models}
  \label{target-model-accuracy-tbl}
  \centering
  \begin{tabular}{lllllll}
    \toprule
    \midrule
    Dataset(Model)  & Train  & Test & Attack Acc & Naive Attack \\
    \midrule
    MNIST (LeNet) & 99.74\% & 99.05\% & 50.04\% & 50.07\% \\
    C-10 (AlexNet) & 91.80\% & 77.22\% & 57.43\% & 57.87\% \\ 
    C-10 (ResNet) & 99.43\% & 93.89\% & 54.11\% & 52.56\% \\ 
    C-10 (DenseNet) & 100.00\% & 95.46\% & 56.05\% & 52.45\% \\ 
    C-100 (AlexNet) & 92.48\% & 43.87\% & 74.02\% & 70.23\% \\
    C-100 (ResNet) & 95.80\% &  74.14\% & 61.70\% & 65.68\% \\
    C-100 (DenseNet) & 99.98\% & 82.83\% & 66.65\% & 71.97\% \\
    I (InceptionV3) & 87.91\% & 79.98\% & 50.03\% & 54.12\% \\
    I (Xception)  & 87.77\% & 80.70\% & 51.18\% & 53.37\% \\
    \bottomrule
  \end{tabular}
\end{table}

\begin{table*}
\scriptsize
  \caption{Membership attack results based on confidence values}
  \label{conf-tbl}
  \centering
  \begin{tabular}{llllll}
    \toprule
    Dataset(Model) & - & Attack Accuracy & Attack FAR & Train Confidence & Test Confidence \\
    \midrule
    & All data & 50.04\% $\pm$ 0.11 & 50.01\% $\pm$ 47.81 & 99.61 $\pm$ 4.57 & 98.90 $\pm$ 9.14      \\
    MNIST & Correctly classified & 49.98\% $\pm$ 0.01 & 50.21\% $\pm$ 48.38 & 99.81 $\pm$ 2.04 & 99.75 $\pm$ 2.51 \\
    & Misclassified & 62.30\% $\pm$ 17.93 & 40.83\% $\pm$ 35.44 & 77.61 $\pm$ 16.05 & 87.09 $\pm$ 15.93   \\
    \midrule
    & All data & 57.43\% $\pm$ 2.59 & 71.4\% $\pm$ 9.29 & 85.26 $\pm$ 23.08 & 72.56 $\pm$ 34.75 \\
    CIFAR-10 (AlexNet) & Correctly classified & 50.54\% $\pm$ 0.82 & 91.89\% $\pm$ 5.43 & 90.77 $\pm$ 13.96 & 89.57 $\pm$ 15.52 \\
    & Misclassified & 52.16\% $\pm$ 2.57 & 5.45\% $\pm$ 5.62 & 60.17 $\pm$ 16.68 & 66.89 $\pm$ 19.68 \\
    \midrule
    & All data & 54.11\% $\pm$ 1.92 & 86.60\% $\pm$ 6.49 & 98.66 $\pm$ 7.03 & 92.74 $\pm$ 22.02 \\
    CIFAR-10 (ResNet) & Correctly classified & 51.81\% $\pm$ 0.84 & 91.63\% $\pm$ 4.05 & 99.08 $\pm$ 4.33 & 97.98 $\pm$ 7.33 \\
    & Misclassified & 60.83\% $\pm$ 19.74 & 0.0\% $\pm$ 0.0 & 66.23 $\pm$ 15.66 & 79.71 $\pm$ 18.61 \\
    \midrule
    & All data & 56.05\% $\pm$ 3.88 & 77.05\% $\pm$ 26.5 & 99.97 $\pm$ 0.49 & 94.78 $\pm$ 19.33 \\
    CIFAR-10 (DenseNet) & Correctly classified & 54.0\% $\pm$ 2.64 & 81.15\% $\pm$ 27.45 & 99.97 $\pm$ 0.29 & 98.77 $\pm$ 5.64 \\
    & Misclassified & 100.0\% $\pm$ 0.0 & 0.0\% $\pm$ 0.0 & - & 82.83 $\pm$ 17.63 \\
    \midrule
    & All data & 74.02\% $\pm$ 8.27 & 38.89\% $\pm$ 16.69 & 84.59 $\pm$ 24.31 & 40.58 $\pm$ 42.09 \\
    CIFAR-100 (AlexNet) & Correctly classified & 55.13\% $\pm$ 7.39 & 83.12\% $\pm$ 14.94 & 89.9 $\pm$ 15.82 & 85.05 $\pm$ 19.95 \\
    & Misclassified & 55.11\% $\pm$ 11.28 & 2.01\% $\pm$ 4.33 & 50.29 $\pm$ 19.45 & 60.96 $\pm$ 23.81 \\
    \midrule
    & All data & 61.7\% $\pm$ 6.55 & 64.45\% $\pm$ 14.81 & 91.14 $\pm$ 18.44 & 70.19 $\pm$ 38.10 \\
    CIFAR-100 (ResNet) & Correctly classified & 53.96\% $\pm$ 5.25 & 83.7\% $\pm$ 11.01 & 94.15 $\pm$ 11.43 & 90.88 $\pm$ 15.74 \\
    & Misclassified & 54.37\% $\pm$ 16.0 & 2.66\% $\pm$ 8.08 & 57.82 $\pm$ 17.69 & 64.3 $\pm$ 21.71 \\
    \midrule
    & All data & 66.65\% $\pm$ 8.36 & 65.0\% $\pm$ 18.16 & 99.95 $\pm$ 1.07 & 79.99 $\pm$ 34.70 \\
    CIFAR-100 (DenseNet) & Correctly classified & 60.58\% $\pm$ 5.98 & 77.17\% $\pm$ 15.28 & 99.96 $\pm$ 0.5 & 94.67 $\pm$ 13.06 \\
    & Misclassified & 98.25\% $\pm$ 9.2 & 0.0\% $\pm$ 0.0 & 65.01\% $\pm$ 11.52 & 67.25\% $\pm$ 24.61 \\
    \midrule
    & All data & 50.03\% $\pm$ 0.28 & 45.96\% $\pm$ 44.27 & 76.03 $\pm$ 25.52 & 68.62 $\pm$ 29.43      \\
    ImageNet (InceptionV3) & Correctly classified & 50.03\% $\pm$ 0.31 & 45.46\% $\pm$ 47.86 & 83.99 $\pm$ 15.55 & 81.85 $\pm$ 16.3 \\
    & Misclassified & 51.5\% $\pm$ 8.57 & 51.69\% $\pm$ 44.69 & 13.57 $\pm$ 11.44 & 10.8 $\pm$ 9.38   \\
    \midrule
    & All data & 51.18\% $\pm$ 3.56 & 50.81\% $\pm$ 44.33 & 73.56 $\pm$ 25.56 & 66.92 $\pm$ 28.58      \\
    ImageNet (Xception) & Correctly classified & 50.72\% $\pm$ 3.57 & 50.42\% $\pm$ 48.80 & 81.24 $\pm$ 16.81 & 79.08 $\pm$ 17.18     \\
    & Misclassified & 51.95\% $\pm$ 19.29 & 52.15\% $\pm$ 44.08 & 13.48 $\pm$ 10.94 & 11.27 $\pm$ 8.94   \\
    \bottomrule
  \end{tabular}
\end{table*}

\begin{figure*}
\def\tabularxcolumn#1{m{#1}}
\begin{tabularx}{\linewidth}{@{}cXX@{}}
\begin{tabular}{ccc}
\subfloat[All samples]{\includegraphics[width=0.3\linewidth]{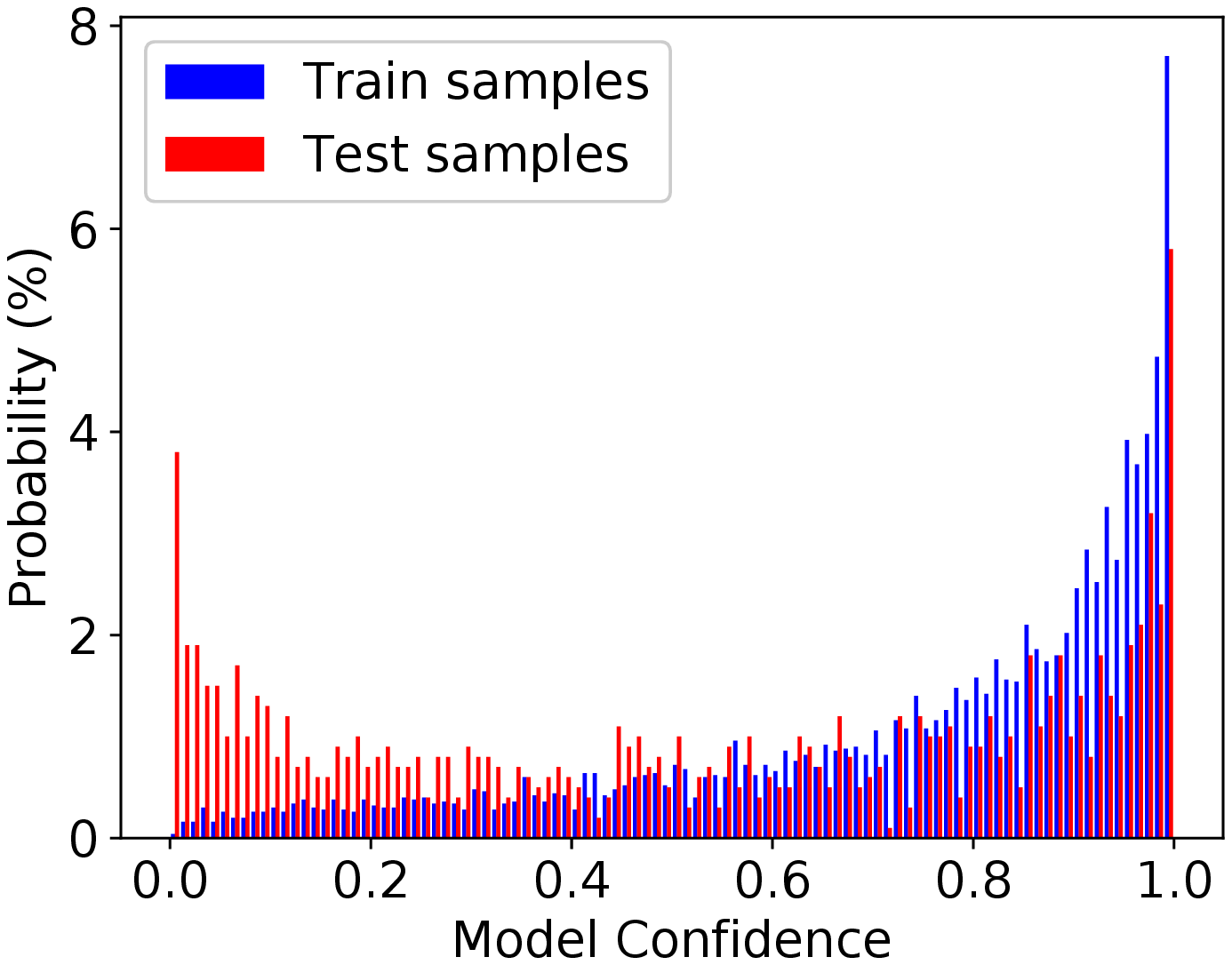}} 
& \subfloat[Correctly classified samples]{\includegraphics[width=0.3\linewidth]{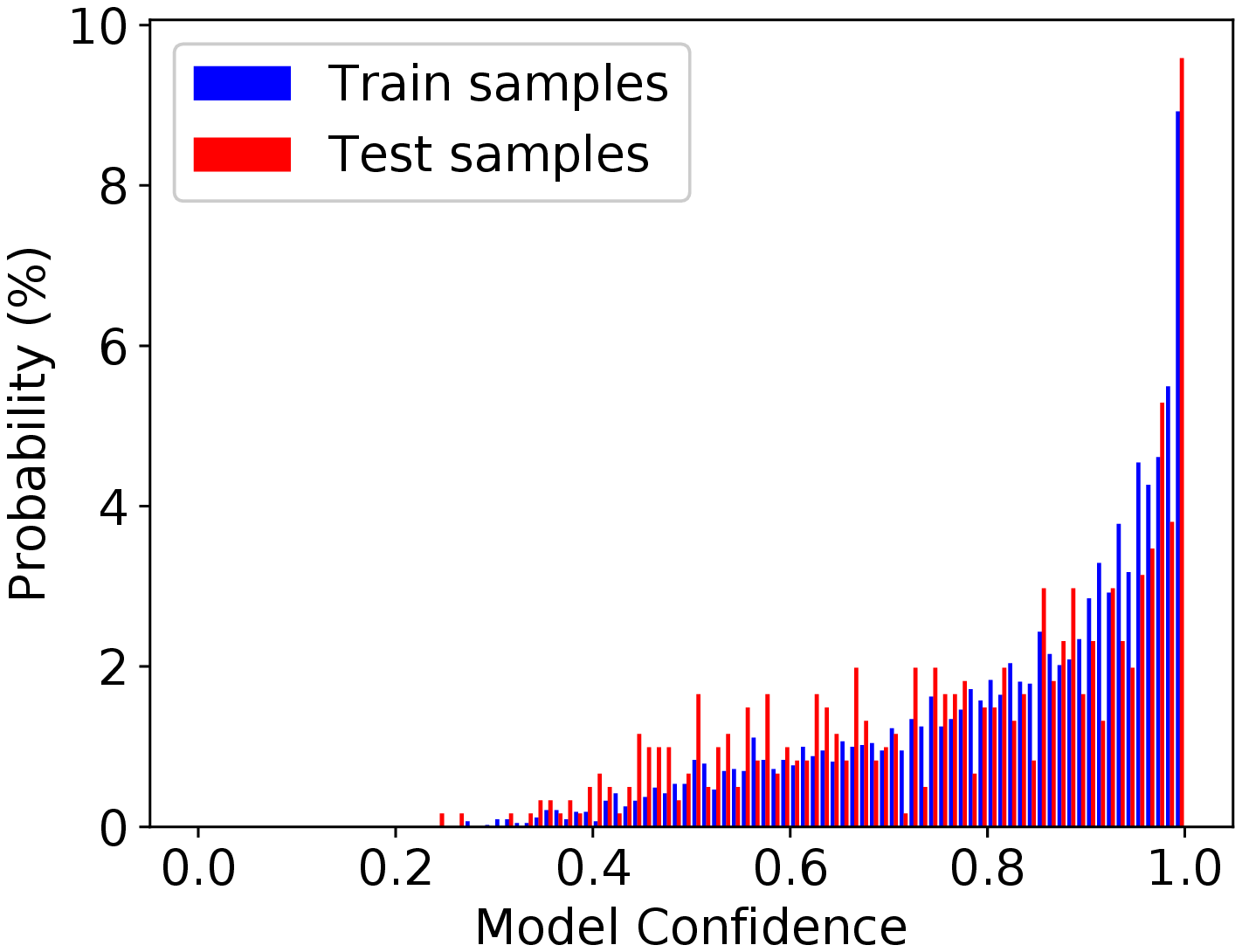}}
& \subfloat[Misclassified samples]{\includegraphics[width=0.3\linewidth]{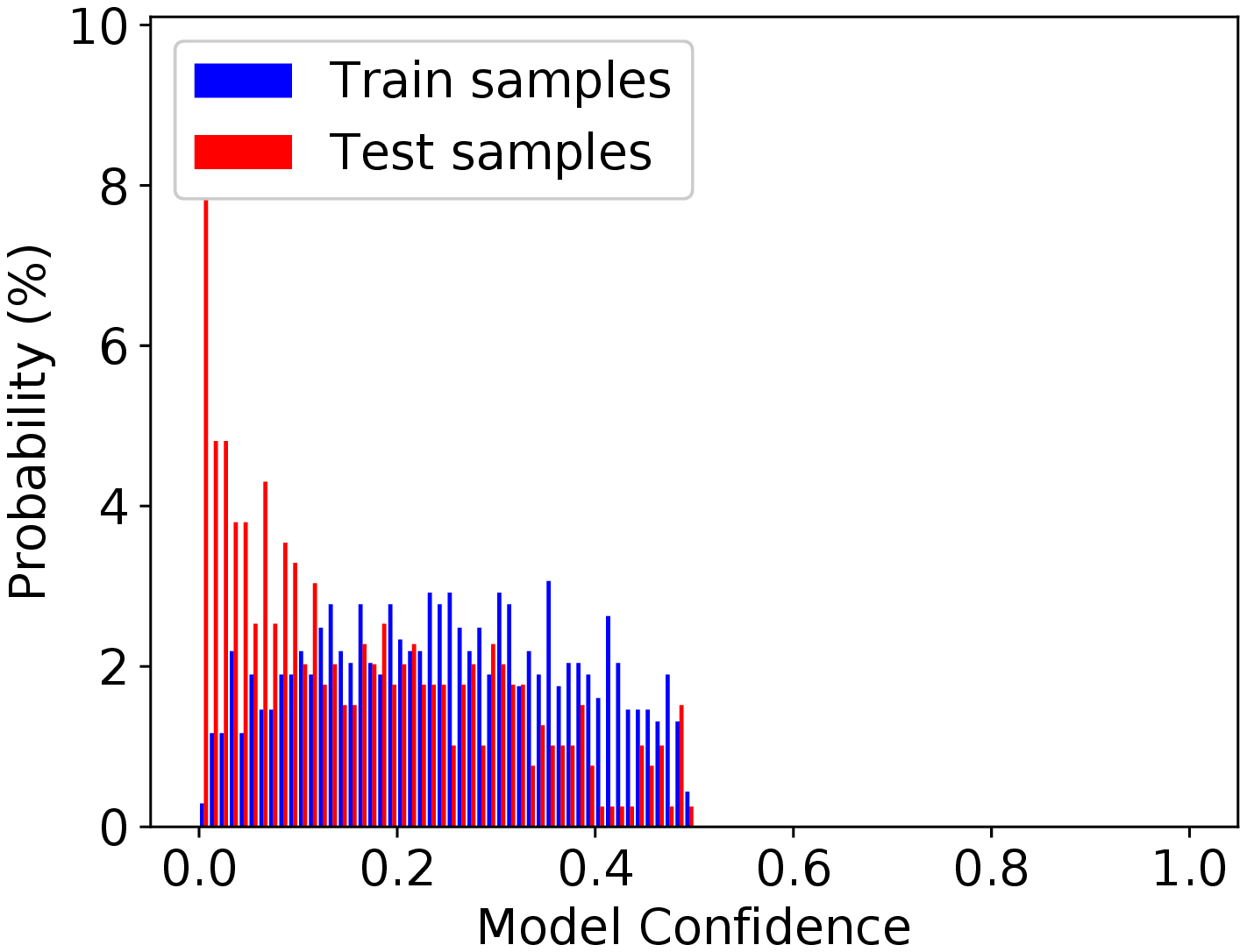}}\\
\end{tabular}
\end{tabularx}
\caption{Distribution of the confidence of the true class for CIFAR-10 (AlexNet) class \#3. Although the distribution of all samples seems to be distinguishable, it is only the manifestation of accuracy gap between train and test which is exploited by naive and other attack. When correctly classified and misclassified samples are depicted separately, sample difference of train and test sets is vividly less distinguishable.}\label{correclty-labeled-conf-dist-fig}
\end{figure*}

\section{Experimental Evidence}
\label{eval}

\paragraph{Target models and datasets} We launch MI attack on various CNN-based models on four image classification tasks: MNIST, CIFAR-10 (C-10), CIFAR-100 (C-100), ImageNet (I). For MNIST, we train a LeNet model. For CIFAR-10 and CIFAR-100, we
use a set of trained models used in \cite{nasr2019comprehensive}\footnote{https://github.com/bearpaw/pytorch-classification}, including AlexNet, ResNet \cite{he2016deep}, and DenseNet \cite{huang2017densely}. For ImageNet, we use pre-trained InceptionV3 \cite{szegedy2016rethinking} and Xception \cite{chollet2017xception} models without any re-training from Keras package\footnote{https://keras.io/api/applications/}. We deliberately choose to launch MI attacks on models trained by others for two reasons: 1) to make it easier for readers to compare the attack with papers that use the same trained models, such as \cite{nasr2019comprehensive}, and 2) to avoid any intentional/unintentional model training practices that bias our results. To the best of our knowledge, the models trained on CIFAR-10 and CIFAR-100, used in \cite{nasr2019comprehensive}, has not adopted the early stopping technique during the training phase. In supplementary material, we train all models on CIFAR-10 and CIFAR-100 from scratch and we show that using early stopping techniques can make MI attacks even harder.

\begin{table*}
\scriptsize
  \caption{MI attack performance based on the output of intermediate layers. C and I represents CIFAR and ImageNet, respectively}
  \label{intermediate-layers-tbl}
  \centering
  \begin{tabular}{lllllllll}
    \toprule
    Dataset (Model) & Layer & \multicolumn{2}{c}{All Data} & \multicolumn{2}{c}{Correctly Classified} & \multicolumn{2}{c}{Misclassified} \\
    \cmidrule(r){3-4}
    \cmidrule(r){5-6}
    \cmidrule(r){7-8}
    -    & - & Accuracy & FAR & Accuracy & FAR \\
    \midrule
    MNIST & -1 & 47.65\% $\pm$ 2.6 & 59.55\% $\pm$ 17.95 & 47.58\% $\pm$ 2.6 & 59.69\% $\pm$ 17.98 & 55.45\% $\pm$ 20.49 & 43.33\% $\pm$ 40.1 \\
    \midrule
    MNIST & -2 & 47.96\% $\pm$ 3.01 & 60.77\% $\pm$ 11.33 & 47.99\% $\pm$ 2.93 & 60.65\% $\pm$ 11.28 & 47.42\% $\pm$ 25.3 & 69.17\% $\pm$ 41.51 \\
    \midrule
    C-10 (AlexNet) & -1 & 55.38\% $\pm$ 2.18 & 47.45\% $\pm$ 9.55 & 51.34\% $\pm$ 2.46 & 58.47\% $\pm$ 12.79 & 55.47\% $\pm$ 3.69 & 14.49\% $\pm$ 9.95 \\
    \midrule
    C-10 (ResNet) & -1 & 53.89\% $\pm$ 2.62 & 45.1\% $\pm$ 16.25 & 52.74\% $\pm$ 2.2 & 47.63\% $\pm$ 17.43 & 60.75\% $\pm$ 20.8 & 5.51\% $\pm$ 7.21 \\
    \midrule
    C-10 (DenseNet) & -1 & 54.4\% $\pm$ 3.63 & 48.7\% $\pm$ 6.99 & 53.12\% $\pm$ 3.03 & 51.28\% $\pm$ 7.54 & 100.0\% $\pm$ 0.0 & 0.0\% $\pm$ 0.0 \\
    \midrule
    C-100 (AlexNet) & -1 & 61.34\% $\pm$ 7.91 & 38.95\% $\pm$ 12.48 & 55.39\% $\pm$ 10.45 & 52.46\% $\pm$ 19.99 & 57.65\% $\pm$ 12.98 & 29.32\% $\pm$ 15.65 \\
    \midrule
    C-100 (ResNet) & -1 & 53.81\% $\pm$ 7.24 & 47.2\% $\pm$ 16.65 & 51.46\% $\pm$ 7.72 & 52.85\% $\pm$ 18.99 & 52.39\% $\pm$ 23.72 & 31.2\% $\pm$ 27.8 \\
    \midrule
    C-100 (DenseNet) & -1 & 64.76\% $\pm$ 9.99 & 37.35\% $\pm$ 14.67 & 61.7\% $\pm$ 9.4 & 43.48\% $\pm$ 14.99 & 92.02\% $\pm$ 20.78 & 38.73\% $\pm$ 31.23 \\
    \midrule
    I (InceptionV3) & -1 & 58.37\% $\pm$ 7.8 & 40.2\% $\pm$ 15.68 & 57.83\% $\pm$ 9.4 & 42.58\% $\pm$ 19.0 & 55.86\% $\pm$ 17.26 & 36.25\% $\pm$ 35.4 \\
    \midrule
    I (Xception) & -1 & 57.44\% $\pm$ 8.59 & 42.3\% $\pm$ 17.08 & 57.35\% $\pm$ 9.36 & 43.92\% $\pm$ 18.61 & 55.52\% $\pm$ 18.65 & 40.38\% $\pm$ 39.49\\
    \bottomrule
  \end{tabular}
\end{table*}

\paragraph{MI attack models}
In most cases, we fit three types of attack models: FC neural network (NN), random forest (RF), and XGBoost. For the NN model, we train a model with 2 hidden layers of size 128 and 64. For RF and XGBoost, we perform a random search over a large set of hyper-parameters.
For ImageNet, we conduct experiments with only 100 classes out of 1000 classes due to the limited computational and time budget. Moreover, we perform random under-sampling of member class and oversampling of non-member class to balance the training dataset on separate experiments. In this paper, we only report the seemingly best MI attack accuracy we achieve over all attack models and hyper-parameters. It is worth mentioning that by under-sampling or over-sampling of training data, or by chainging the decision threshold of the MI attack, we could decrease FAR at the cost of accuracy. In any case, we could not find a good MI attack model with relatively high accuracy and low FAR. 
The input of attack models varies which is described in each following subsection. The accuracy of target and MI attack models are shown in Table \ref{target-model-accuracy-tbl}. Note that even the best MI attack models can barely outperform the naive attack. In the following sections, we show that separating correctly classified and misclassified samples, and reporting accuracy and FAR gives more insight on the performance of MI attacks.

\subsection{Confidence Values}
Confidence values have been extensively used for MI attacks. As shown in Table \ref{conf-tbl}, the MI attacks are more successful on inferring membership of misclassified samples, which often consist a small portion of training samples. Interestingly, the state-of-the-art target models on ImageNet does not even leak membership status of misclassified samples. The best attack performance on correctly classified samples is observed on DenseNet model trained on CIFAR-100, which is $60.58\%$ that still suffers from very high FAR ($77.17\%$). Note that the MI attacks on misclassified samples of DenseNet model may not be meaningful because there are no misclassified samples in the training set of CIFAR-10 and there are only 10 misclassified samples in the training set of CIFAR-100.

To better understand why MI attacks fail, it is better to investigate the average confidence value of target models, shown in the fifth and sixth columns of Table \ref{conf-tbl}. As shown, the average confidence values of train samples (members) are often close to the test samples (non-members). MI attacks are only partially successful when average confidence values between train and test samples are far apart and the standard deviation is low. As shown in Figure \ref{correclty-labeled-conf-dist-fig}, by separating correctly classified samples and misclassified sample, we can observe that sample distribution is very close, particularly for correctly classifies samples.

\subsection{Output of Intermediate Layers}
The attack accuracy based on the output of intermediate layers are shown in Table \ref{intermediate-layers-tbl}. We only launch an attack on the output of FC or flattened layers. The layer column shows the number of layers we go back from the Softmax layer. Only the MI attack on ImageNet is more successful in terms of both accuracy and FAR. Nevertheless, the FAR is still high and accuracy is not considerably better than a random guess.

\begin{table*}
\scriptsize
  \caption{Performance of attack models based on the distance to the decision boundary.}
  \label{distance-tbl}
  \centering
  \begin{tabular}{llllllllll}
    \toprule
    Dataset (Model) & \multicolumn{4}{c}{Correctly Classified} & \multicolumn{4}{c}{Misclassified} \\
    \cmidrule(r){2-5}
    \cmidrule(r){6-9}
    -     & \multicolumn{2}{c}{MI Attack} & \multicolumn{2}{c}{Average Distance} & \multicolumn{2}{c}{MI Attack} & \multicolumn{2}{c}{Average Distance} \\
    \cmidrule(r){2-3}
    \cmidrule(r){4-5}
    \cmidrule(r){6-7}
    \cmidrule(r){8-9}
    -     & Accuracy & FAR & Train & Test & Accuracy & FAR & Train & Test \\
    \midrule
    MNIST & 49.86\% $\pm$ 6.0 & 52.97\% $\pm$ 8.2 & 1.372 $\pm$ 0.40 & 1.371 $\pm$ 0.45 & 49.81\% $\pm$ 9.0 & 48.3\% $\pm$ 24.0 & .0103 $\pm$ .0016 & .0108 $\pm$ .0024 &  \\
    \midrule
    C-10 (AlexNet) & 52.36\% $\pm$ 4.1 & 53.19\% $\pm$ 17 & 51.59 $\pm$ 15.2 & 50.31 $\pm$ 14.7 & 52.42\% $\pm$ 4.9 & 49.53\% $\pm$ 7.1 & 48.84 $\pm$ 15.8 & 50.5 $\pm$ 16.0 \\
    \midrule
    C-10 (ResNet) & 51.05\% $\pm$ 5.5 & 49.57\% $\pm$ 8.2 & 51.3 $\pm$ 15.2 & 50.3 $\pm$ 14.7 & 46.05\% $\pm$ 4.8 & 60.49\% $\pm$ 8.6 & 51.2 $\pm$ 15.2 & 50.5 $\pm$ 16.0 \\
    \midrule
    C-10 (DenseNet) & 50.14\% $\pm$ 5.9 & 50.17\% $\pm$ 7.8 & 51.4 $\pm$ 15.3 & 50.2 $\pm$ 14.6 & 100.0\% $\pm$ 0.0 & 0.0\% $\pm$ 0.0 & -  & 51.2 $\pm$ 16.4\\
    \midrule
    C-100 (AlexNet) & 50.45\% $\pm$ 9.1 & 47.77\% $\pm$ 6.3 & 53.6 $\pm$ 16.3 & 54.3 $\pm$ 15.1 & 53.78\% $\pm$ 11.6 & 45.94\% $\pm$ 17.7 & 48.2 $\pm$ 16.5 & 52.6 $\pm$ 16.6 \\
    \midrule
    C-100 (ResNet) & 49.08\% $\pm$ 6.4 & 49.28\% $\pm$ 6.8 & 52.6 $\pm$ 16.4 & 53.3 $\pm$ 15.9 & 47.24\% $\pm$ 12.4 & 52.05\% $\pm$ 21.4 & 51.9 $\pm$ 16.6 & 51.6 $\pm$ 16.8 \\
    \midrule
    C-100 (DenseNet) & 49.74\% $\pm$ 6.41 & 48.57\% $\pm$ 7.04 & 52.8 $\pm$ 16.3 & 53.3 $\pm$ 16.0 & 98.25\% $\pm$ 9.2 & 0.0\% $\pm$ 0.0 & 51.5 $\pm$ 10.4 & 56.8 $\pm$ 16.5 \\
    \midrule
    I (InceptionV3) & 51.75\% $\pm$ 8.2 & 50.43\% $\pm$ 26.3 & 212.4 $\pm$ 64.8 & 218.6 $\pm$ 63.0 & 44.95\% $\pm$ 11.9 & 46.31\% $\pm$ 18.8 & 215.8 $\pm$ 67.0 & 215.1 $\pm$ 61.5 \\
    \midrule
    I (Xception) & 53.01\% $\pm$ 9.5 & 46.1\% $\pm$ 23.7 & 214.5 $\pm$ 64.6 & 216.2 $\pm$ 61.9 & 49.29\% $\pm$ 13.5 & 44.88\% $\pm$ 17.1 & 212.4 $\pm$ 67.0 & 214.7 $\pm$ 64.0 \\
    \bottomrule
  \end{tabular}
\end{table*}

\begin{table*}
\scriptsize
  \caption{Performance of attack models based on gradient norms with respect to input (x) and weights (w)}
  \label{tbl-gradient-norm}
  \centering
  \begin{tabular}{llllllllll}
    \toprule
    Dataset (Model) & \multicolumn{4}{c}{Correctly Classified} & \multicolumn{4}{c}{Misclassified} \\
    \cmidrule(r){2-5}
    \cmidrule(r){6-9}
    -     & \multicolumn{2}{c}{Grad w.r.t w} & \multicolumn{2}{c}{Grad w.r.t x} & \multicolumn{2}{c}{Grad w.r.t w} & \multicolumn{2}{c}{Grad w.r.t x} \\
    \cmidrule(r){2-3}
    \cmidrule(r){4-5}
    \cmidrule(r){6-7}
    \cmidrule(r){8-9}
    -     & Accuracy & FAR & Accuracy & FAR & Accuracy & FAR & Accuracy & FAR \\
    \midrule
    MNIST & 52.06\% $\pm$ 3.7 & 42.75\% $\pm$ 28.5 & 53.19\% $\pm$ 3.5 & 34.5\% $\pm$ 23.7 & 57.84\% $\pm$ 26.8 & 38.48\% $\pm$ 20.0 & 52.02\% $\pm$ 22.2 & 41.79\% $\pm$ 16.8\\
    \midrule
    C-10 (AlexNet) & 51.94\% $\pm$ 4.1 & 41.38\% $\pm$ 7.9 & 51.81\% $\pm$ 4.4 & 39.38\% $\pm$ 10.3 & 61.36\% $\pm$ 6.6 & 39.27\% $\pm$ 8.2 & 58.69\% $\pm$ 5.4 & 35.92\% $\pm$ 7.2\\
    \midrule
    C-10 (ResNet) & 52.75\% $\pm$ 3.0 & 21.75\% $\pm$ 12.2 & 49.88\% $\pm$ 3.8 & 36.25\% $\pm$ 13.9 & 50.85\% $\pm$ 10.6 & 66.49\% $\pm$ 26.8 & 50.26\% $\pm$ 16.1 & 50.36\% $\pm$ 31.3\\
    \midrule
    C-10 (DenseNet) & 54.88\% $\pm$ 2.6 & 16.38\% $\pm$ 5.4 & 54.37\% $\pm$ 3.1 & 13.25\% $\pm$ 8.3 & 
    100.0\% $\pm$ 0.0 & 0.0\% $\pm$ 0.0 & 100.0\% $\pm$ 0.0 & 0.0\% $\pm$ 0.0\\
    \midrule
    C-100 (AlexNet) & 57.71\% $\pm$ 9.6 & 31.45\% $\pm$ 14.6 & 56.65\% $\pm$ 11.5 & 31.46\% $\pm$ 9.9 & 67.32\% $\pm$ 11.3 & 34.51\% $\pm$ 19.9 & 59.12\% $\pm$ 13.8 & 38.52\% $\pm$ 23.5\\
    \midrule
    C-100 (ResNet) & 52.98\% $\pm$ 6.2 & 37.72\% $\pm$ 21.1 & 54.31\% $\pm$ 6.9 & 29.71\% $\pm$ 11.3 & 55.4\% $\pm$ 17.9 & 54.87\% $\pm$ 34.6 & 61.38\% $\pm$ 17.3 & 36.58\% $\pm$ 27.1 \\
    \midrule
    C-100 (DenseNet) & 69.7\% $\pm$ 7.8 & 8.03\% $\pm$ 6.2 & 70.22\% $\pm$ 7.4 & 6.12\% $\pm$ 3.6 & 98.25\% $\pm$ 9.2 & 0.0\% $\pm$ 0.0 & 98.25\% $\pm$ 9.2 & 0.0 $\pm$ 0.0\\
    \midrule
    I (InceptionV3) & 49.25\% $\pm$ 11.7 & 42.5\% $\pm$ 19.0 & 52.1\% $\pm$ 9.6 & 48.91\% $\pm$ 14.7 & 58.48\% $\pm$ 15.3 & 30.8\% $\pm$ 20.3 & 50.55\% $\pm$ 16.5 & 42.75\% $\pm$ 17.3\\
    \midrule
    I (Xception) & 53.48\% $\pm$ 9.0 & 36.53\% $\pm$ 19.3 & 53.26\% $\pm$ 11.4 & 48.05\% $\pm$ 14.2 & 47.89\% $\pm$ 18.2 & 43.69\% $\pm$ 17.7 & 49.65\% $\pm$ 15.6 & 41.25\% $\pm$ 17.4 \\
    \bottomrule
  \end{tabular}
\end{table*}

\subsection{Distance to the Boundary}
Since the distance to the decision boundary is one-dimensional, we only fit a logistic regression to the samples. As shown in Table \ref{distance-tbl}, all MI attacks fail. By looking at the average distance and their standard deviation, it is evident that the distance to boundary is not a distinguishable feature for membership inference. Note that finding the exact distance to the boundary is a computationally heavy task for high dimensional data. What is clear in Table \ref{distance-tbl} is that the FGM-based approximation of distance to boundary does not provide a distinguishable feature. A more accurate approach may reveal more membership information.

\subsection{Gradient Norm}
Table \ref{tbl-gradient-norm} shows the performance of MI attack models based on gradient norms. For each case, we fit a logistic regression to the 7 norms introduced in Section \ref{sec-methodology}. Except for ImageNet, all MI attacks on correctly classified samples achieve almost the same accuracy while having a lower FAR, in comparison with MI attacks based on confidence values (Table \ref{conf-tbl}). Gradient information of misclassified samples leak less membership information than confidence values.

\section{Conclusion}
In this paper, we show that commonly-used MI attacks based on confidence values of deep models are not as reliable as it has been reported before. By reporting accuracy and FAR together, we show that MI attacks that achieve higher accuracy suffers from higher FAR. Previous MI attacks extensively rely on confidence values of the target model for membership inference. We show that such a membership inference in general is a difficult task because the distribution of confidence values are similar for member and non-member samples. We report the attack accuracy on correctly classified samples and misclassified samples separately to show that misclassified samples slightly leak more information for membership inference. Additionally, we analyze several other features of input samples, including the distance to the decision boundary and gradient norms, to further illustrate the difficult nature of reliable membership inference attack on deep models. In summary, we find that naturally trained deep models often behave similarly across training and test samples and, hence, an accurate membership inference attack on all training samples in practice is a difficult and inaccurate task under current attack models unless a new revolutionary approach is introduced.

\textbf{Acknowledgement.} This work was supported by the National Science Foundation (NSF) under Grant CNS-1547461, Grant CNS-1718901, and Grant IIS-1838207


{
\small
\bibliographystyle{ieee_fullname}
\bibliography{ref}
}

\section{Supplementary Material}
\subsection{Effect of Overfitting}
\label{apx-cifar10-depth}

\begin{figure*}[h]
\def\tabularxcolumn#1{m{#1}}
\begin{tabularx}{\linewidth}{@{}cXX@{}}
\begin{tabular}{ccc}
\subfloat[All dataset]{\includegraphics[width=0.3\linewidth]{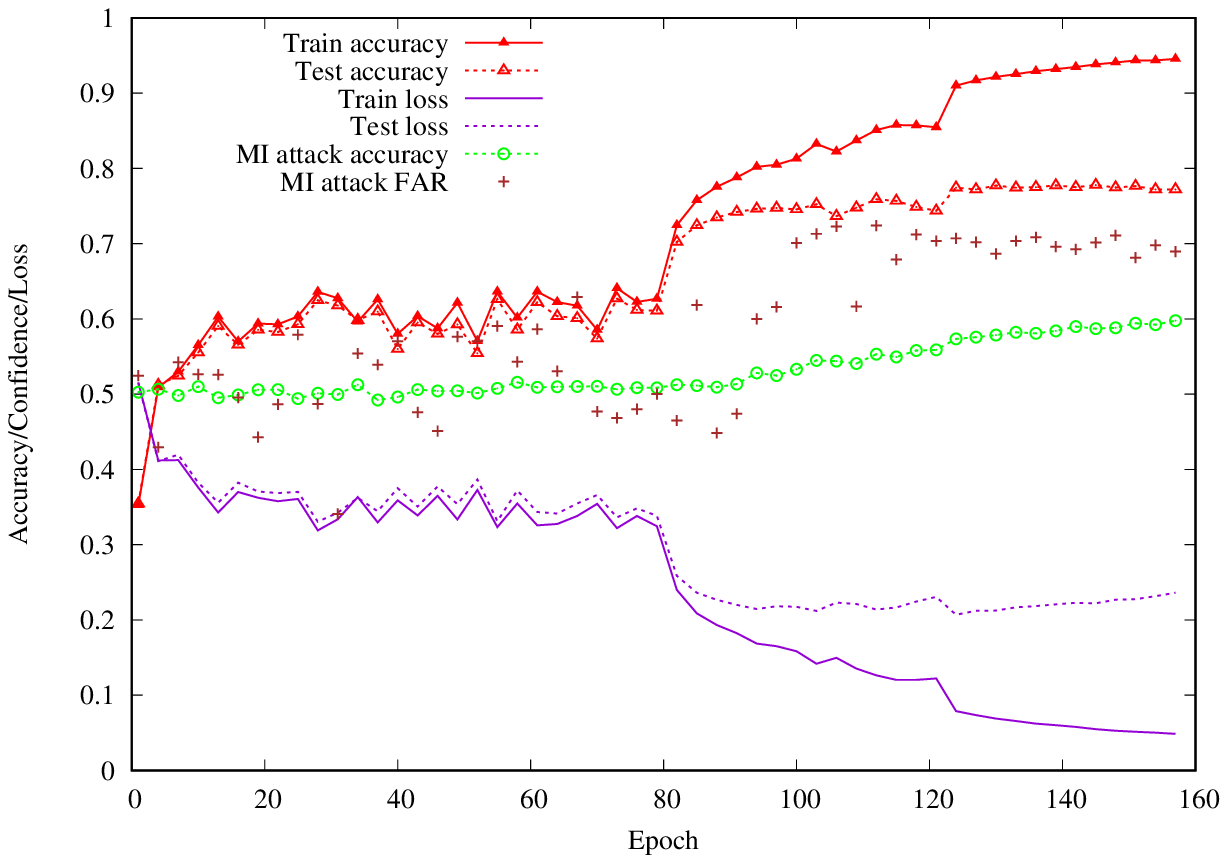}} 
& \subfloat[Correctly classified samples]{\includegraphics[width=0.3\linewidth]{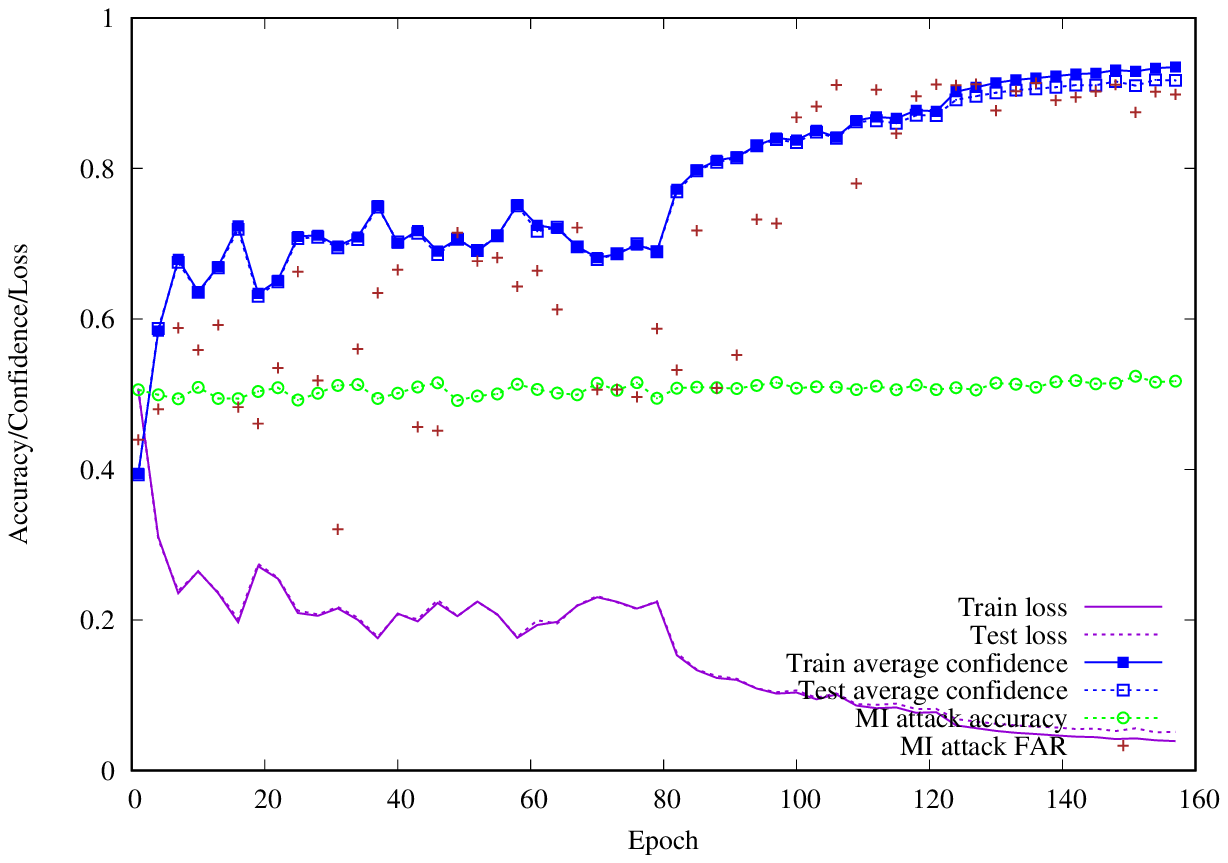}}
& \subfloat[Mis-classified samples]{\includegraphics[width=0.3\linewidth]{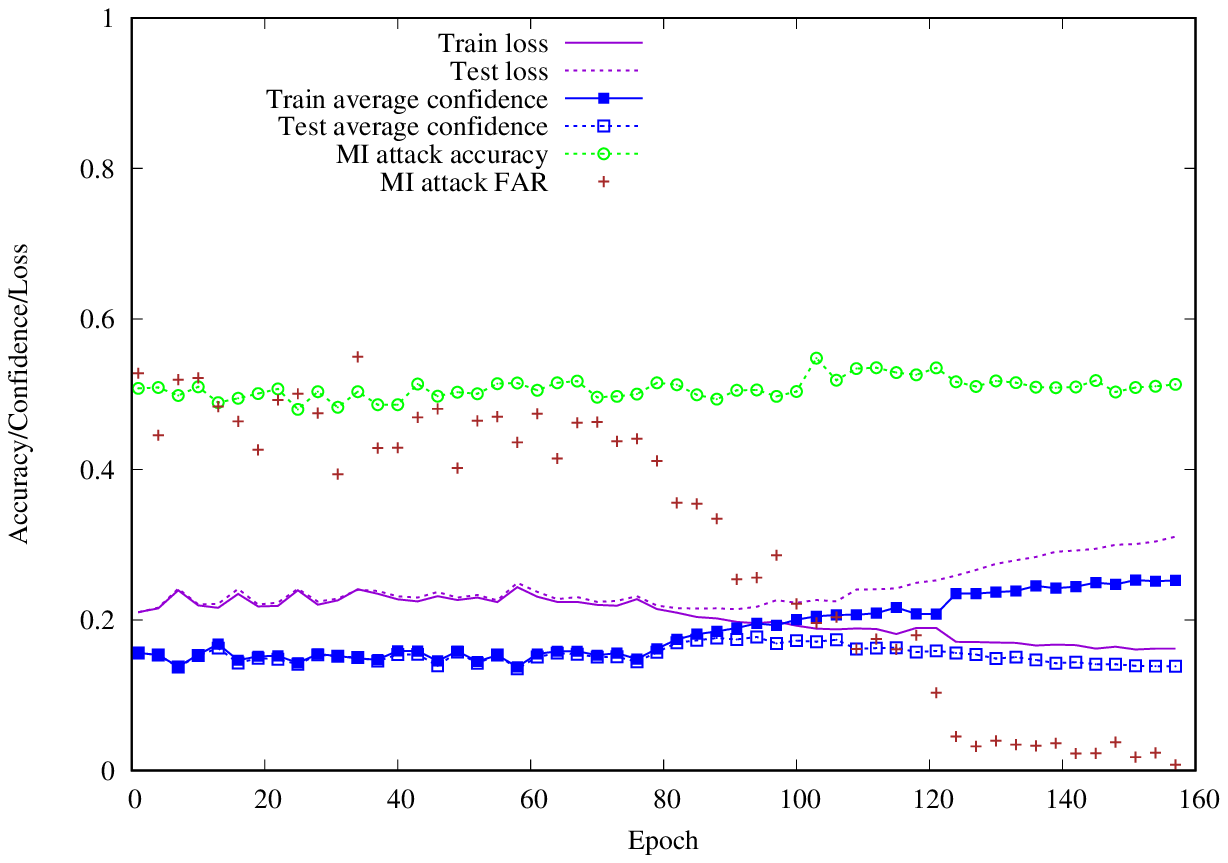}}\\
\end{tabular}
\end{tabularx}
\caption{Training progress and MI attack on CIFAR-10 for AlexNet model}
\label{cifar10-train-progress-alexnet}
\end{figure*}

\begin{figure*}[h]
\def\tabularxcolumn#1{m{#1}}
\begin{tabularx}{\linewidth}{@{}cXX@{}}
\begin{tabular}{ccc}
\subfloat[All dataset]{\includegraphics[width=0.3\linewidth]{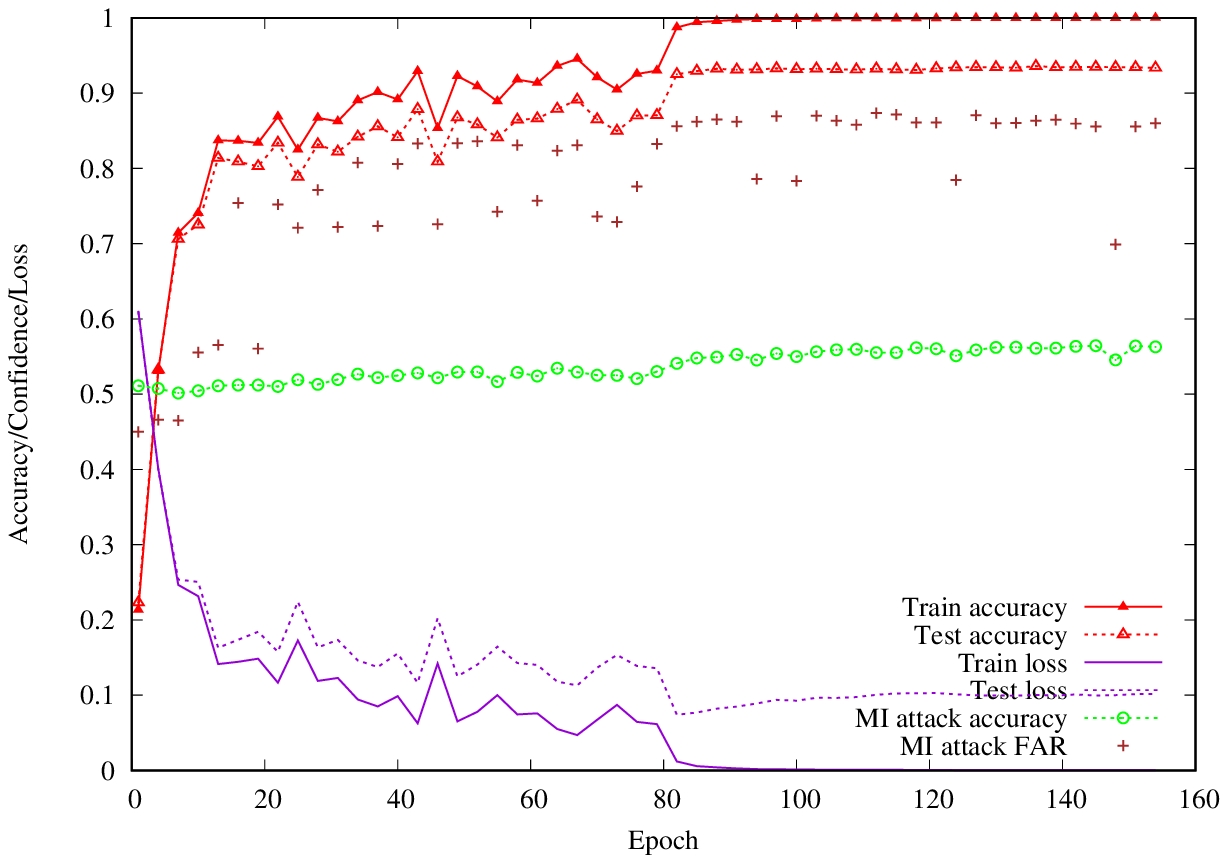}} 
& \subfloat[Correctly classified samples]{\includegraphics[width=0.3\linewidth]{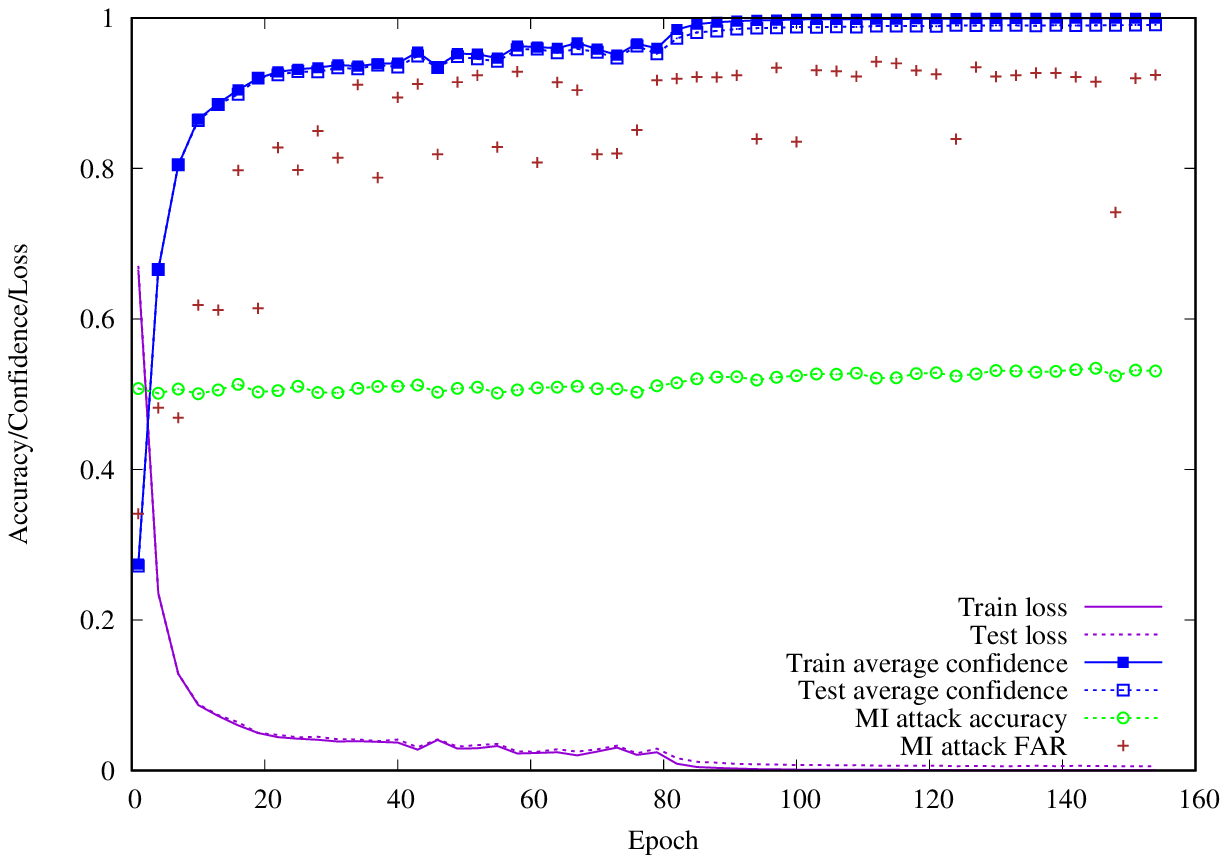}}
& \subfloat[Mis-classified samples]{\includegraphics[width=0.3\linewidth]{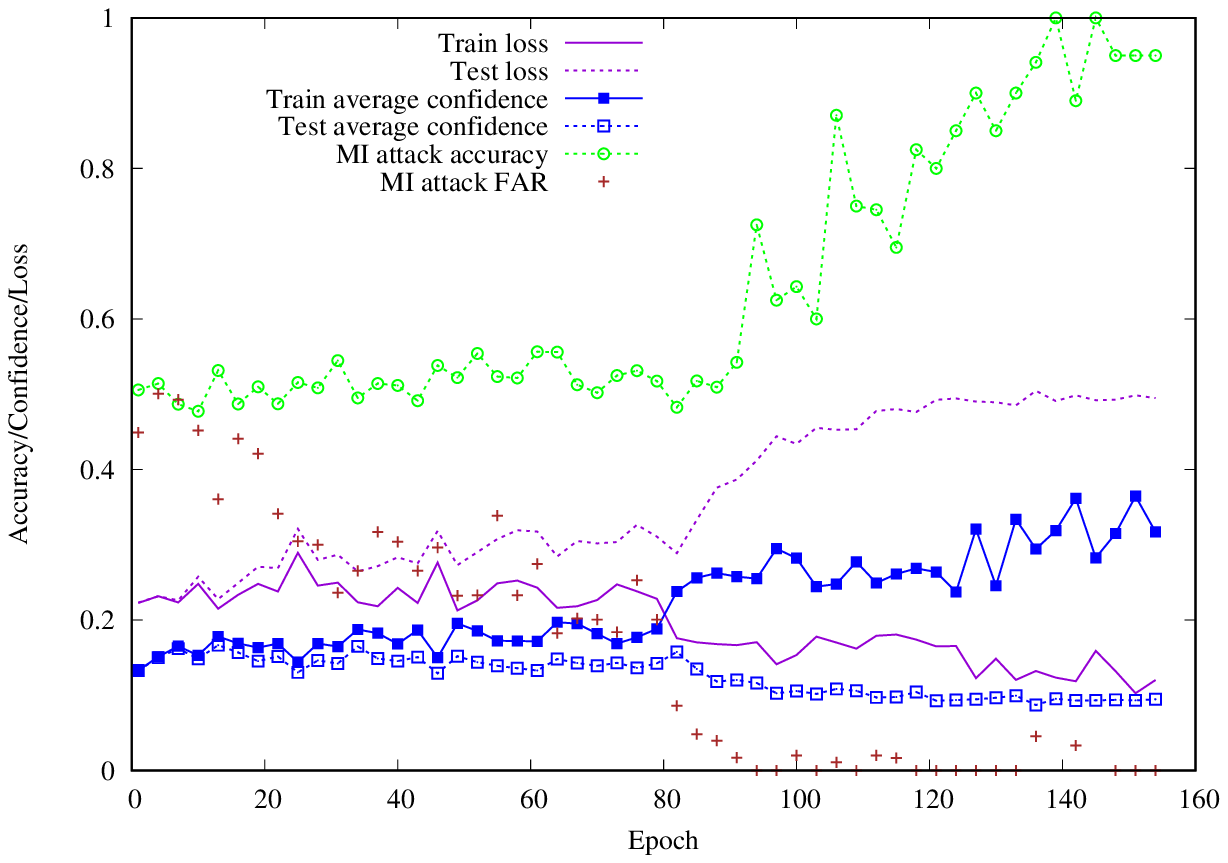}}\\
\end{tabular}
\end{tabularx}
\caption{Training progress and MI attack on CIFAR-10 for ResNet model}
\label{cifar10-train-progress-resnet}
\end{figure*}

\begin{figure*}[h]
\def\tabularxcolumn#1{m{#1}}
\begin{tabularx}{\linewidth}{@{}cXX@{}}
\begin{tabular}{ccc}
\subfloat[All dataset]{\includegraphics[width=0.3\linewidth]{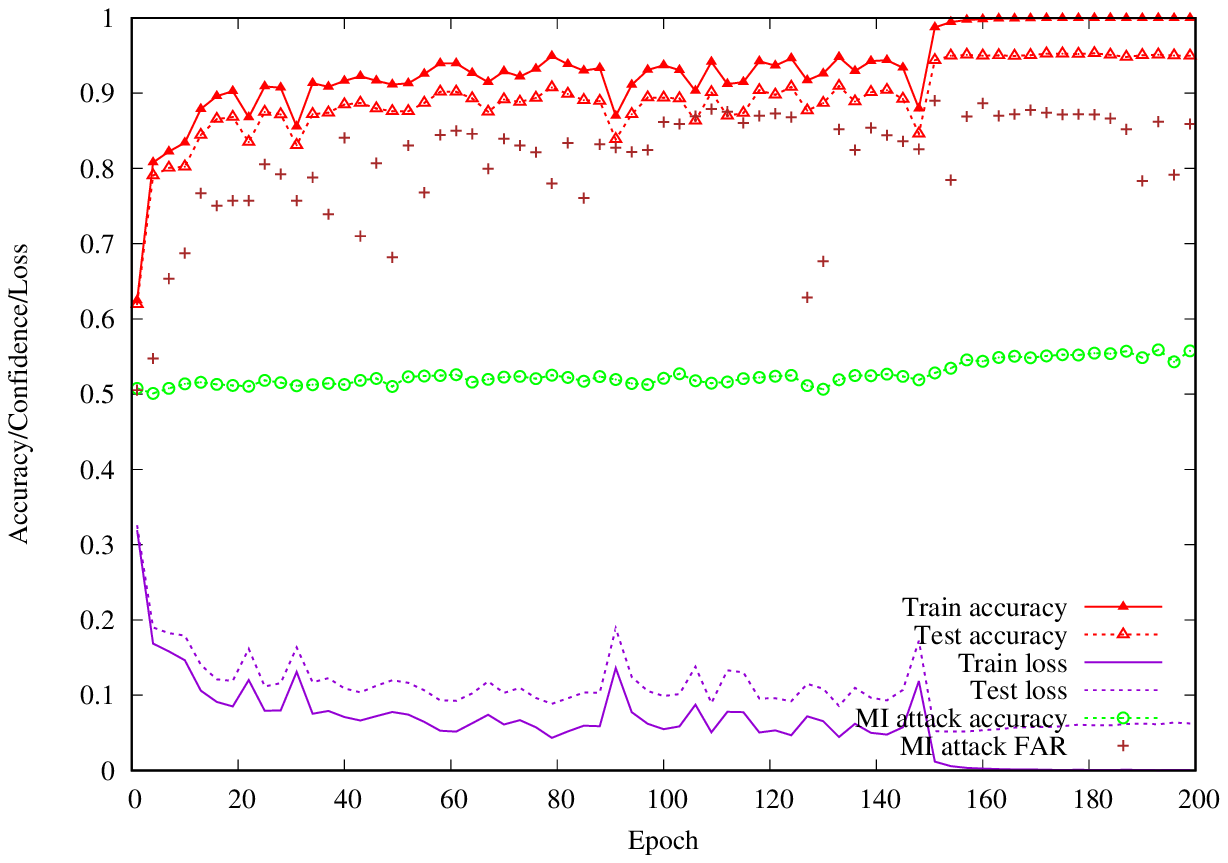}} 
& \subfloat[Correctly classified samples]{\includegraphics[width=0.3\linewidth]{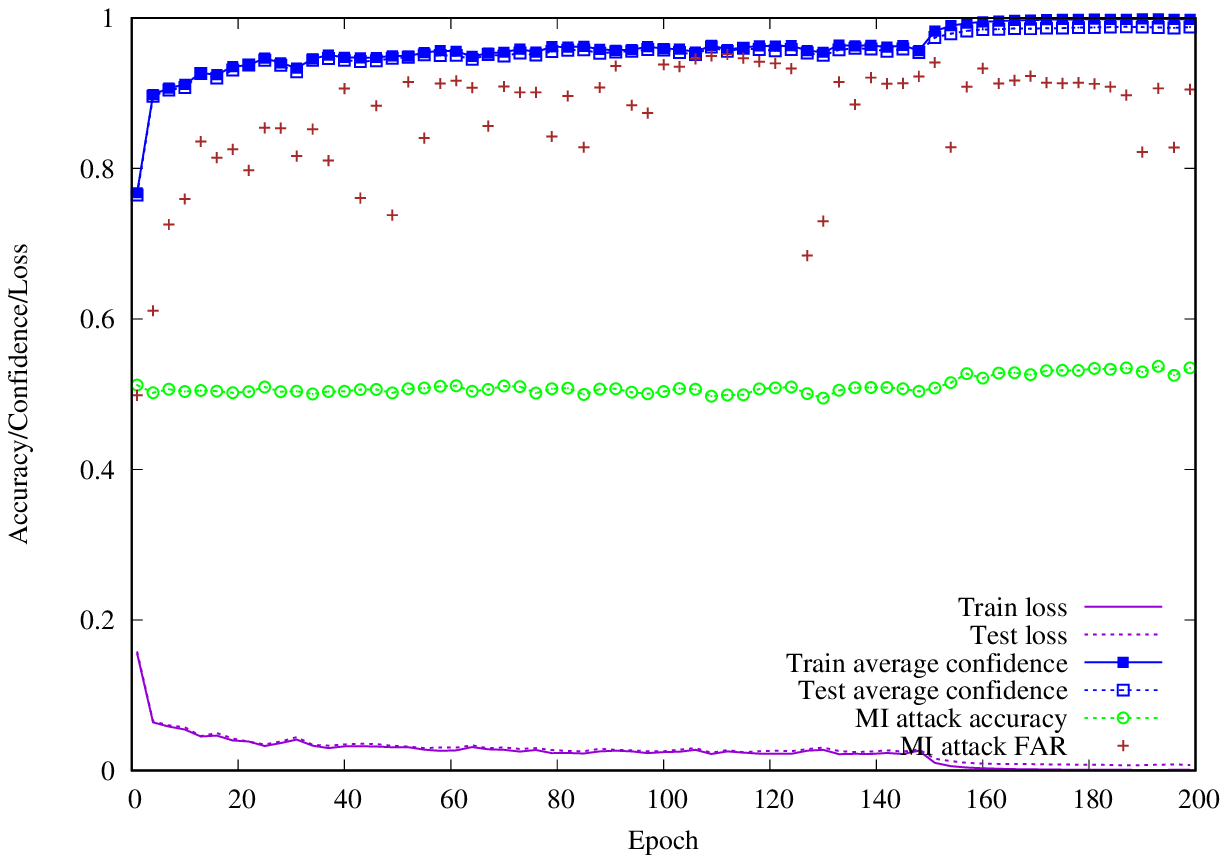}}
& \subfloat[Mis-classified samples]{\includegraphics[width=0.3\linewidth]{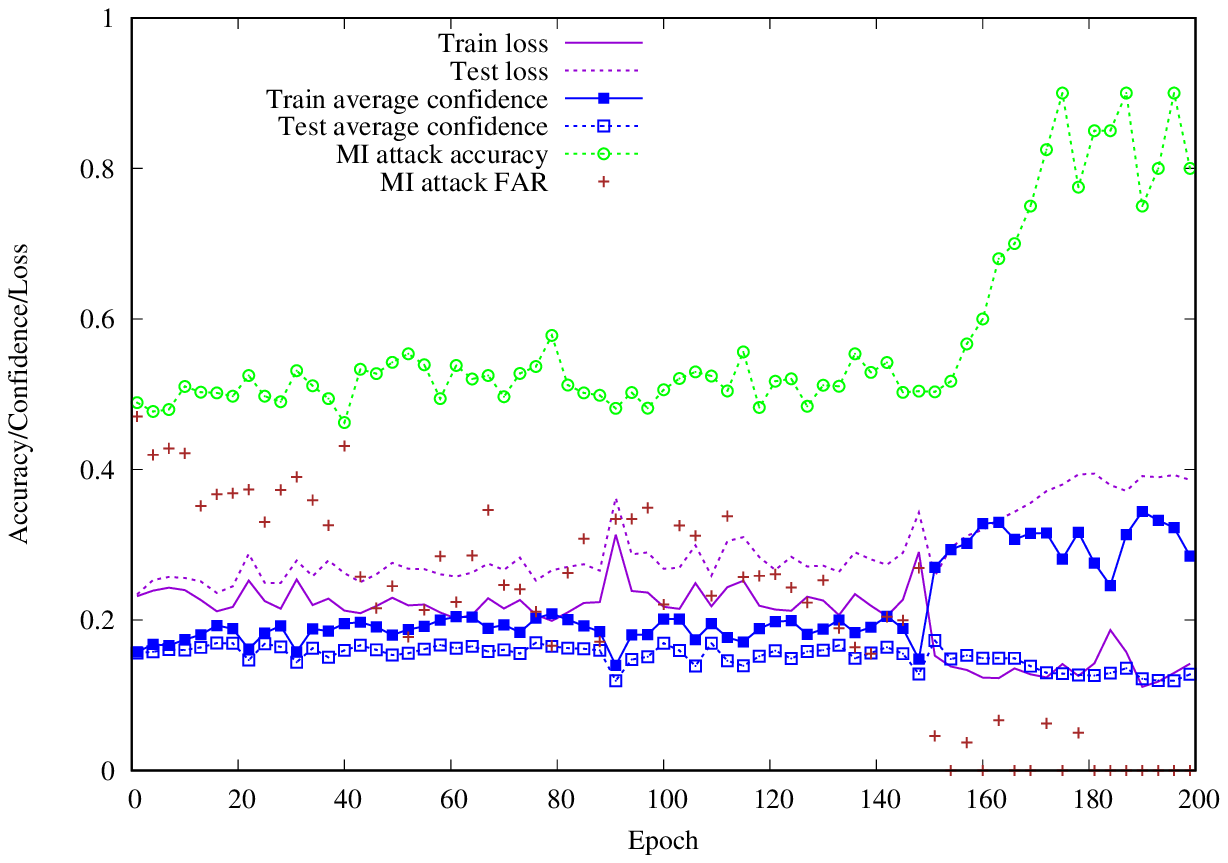}}\\
\end{tabular}
\end{tabularx}
\caption{Training progress and MI attack on CIFAR-10 for DenseNet model}
\label{cifar10-train-progress-densenet}
\end{figure*}

\begin{figure*}[h]
\def\tabularxcolumn#1{m{#1}}
\begin{tabularx}{\linewidth}{@{}cXX@{}}
\begin{tabular}{ccc}
\subfloat[All dataset]{\includegraphics[width=0.3\linewidth]{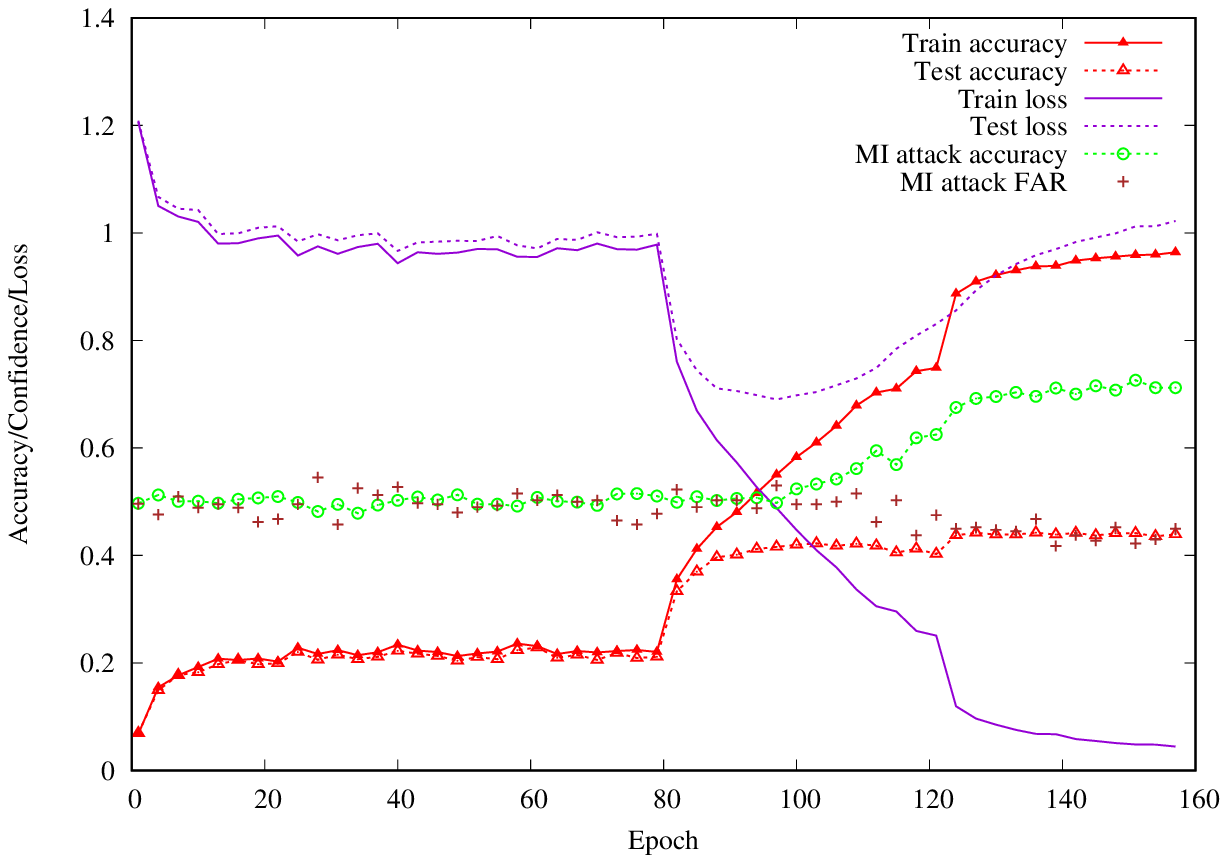}} 
& \subfloat[Correctly classified samples]{\includegraphics[width=0.3\linewidth]{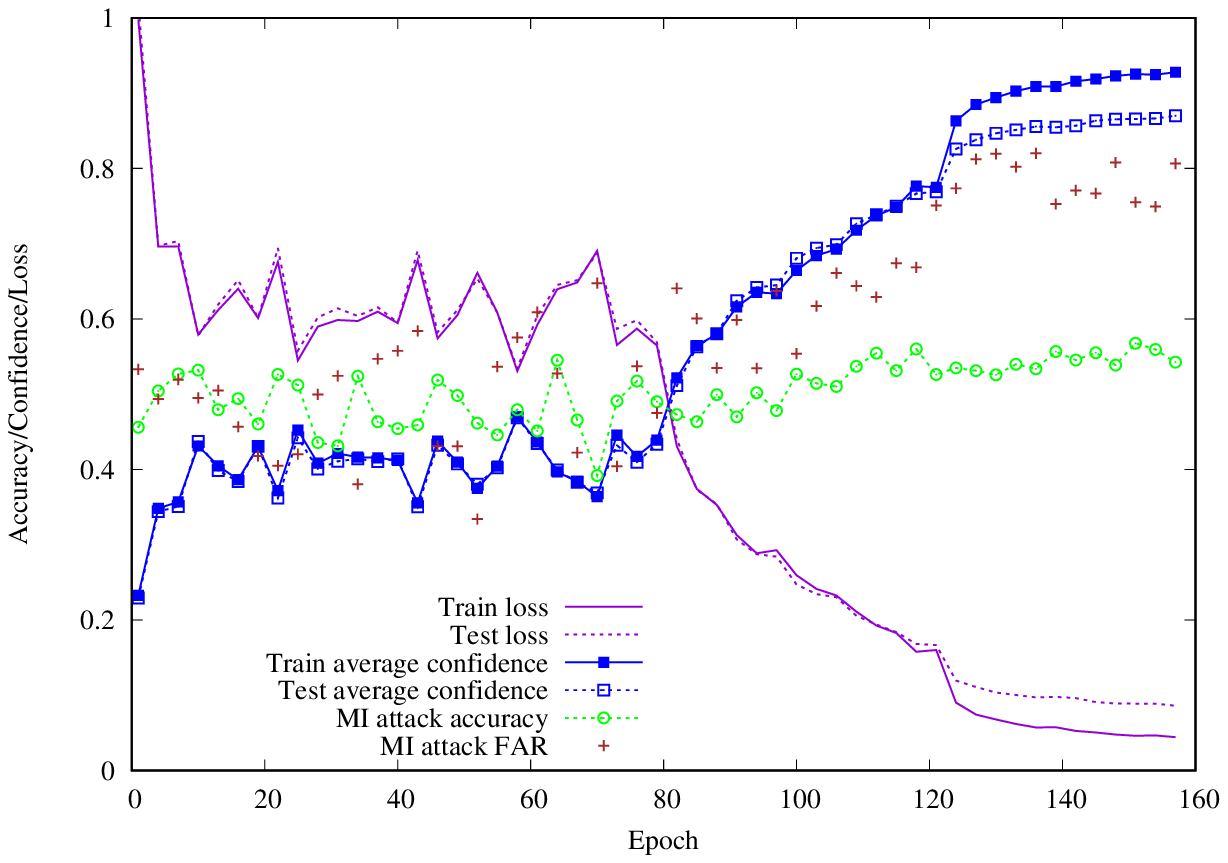}}
& \subfloat[Mis-classified samples]{\includegraphics[width=0.3\linewidth]{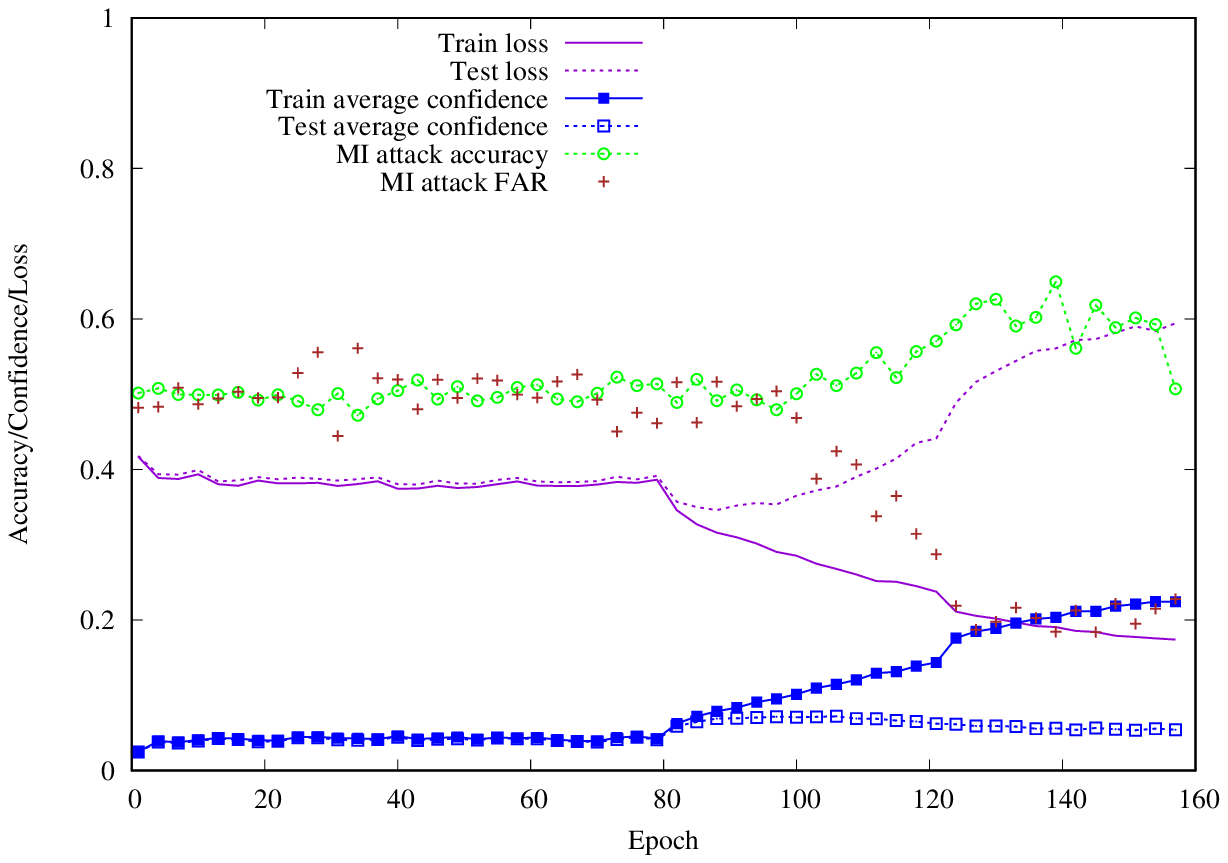}}\\
\end{tabular}
\end{tabularx}
\caption{Training progress and MI attack on CIFAR-100 for AlexNet model}
\label{cifar100-train-progress-alexnet}
\end{figure*}

\begin{figure*}[h]
\def\tabularxcolumn#1{m{#1}}
\begin{tabularx}{\linewidth}{@{}cXX@{}}
\begin{tabular}{ccc}
\subfloat[All dataset]{\includegraphics[width=0.3\linewidth]{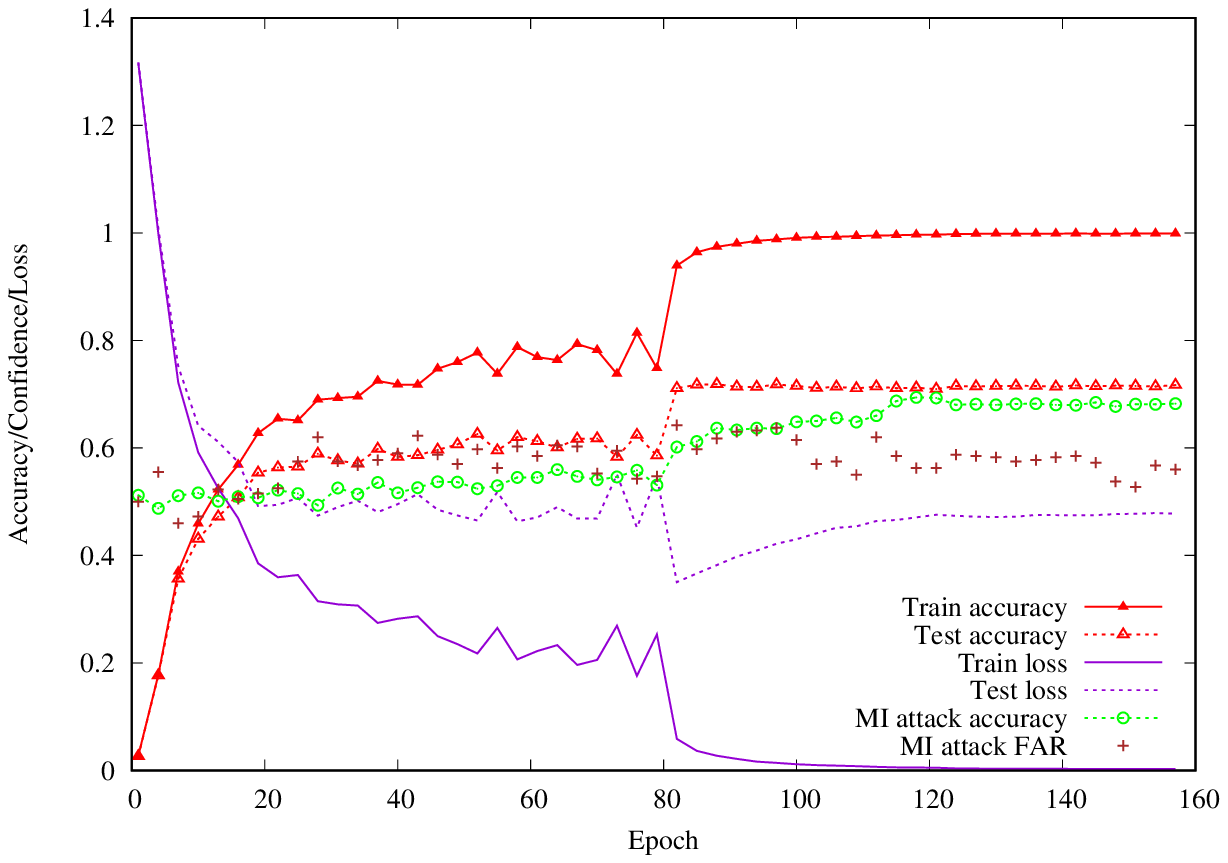}} 
& \subfloat[Correctly classified samples]{\includegraphics[width=0.3\linewidth]{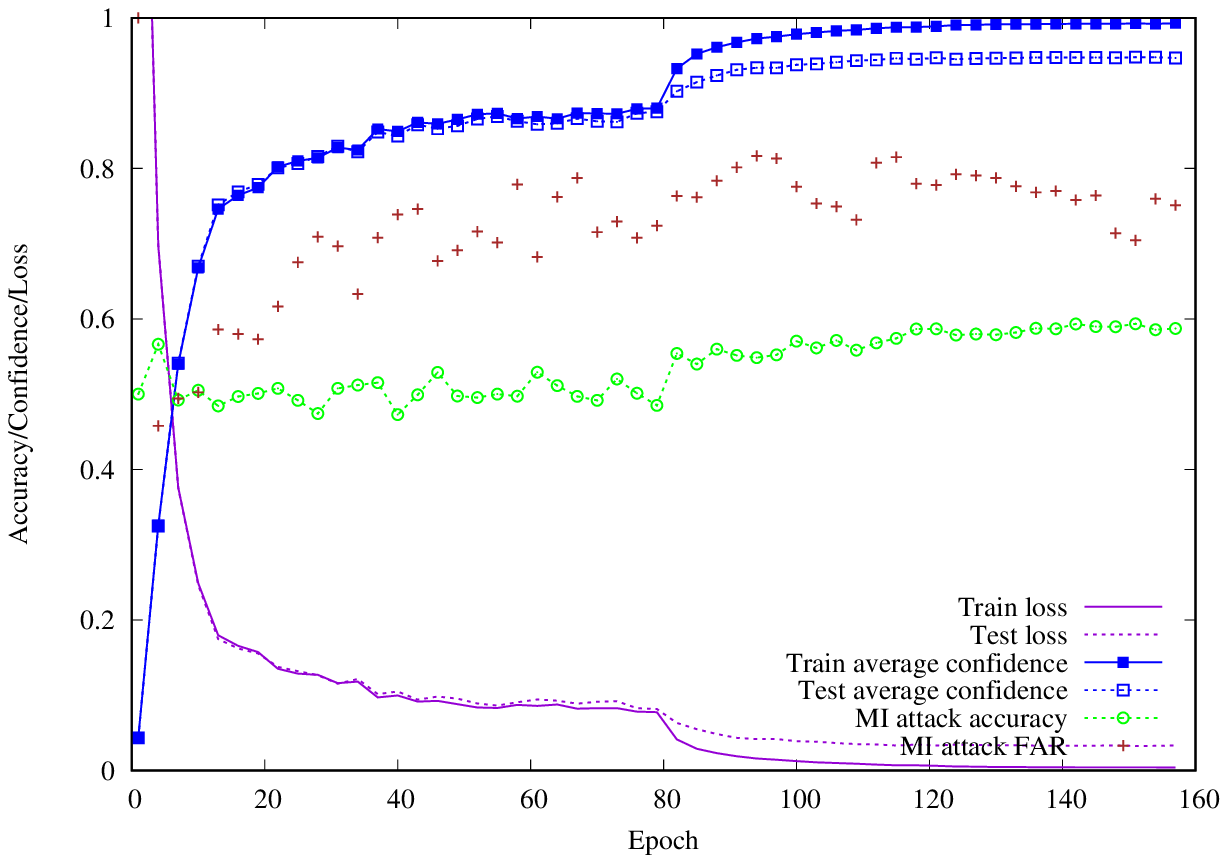}}
& \subfloat[Mis-classified samples]{\includegraphics[width=0.3\linewidth]{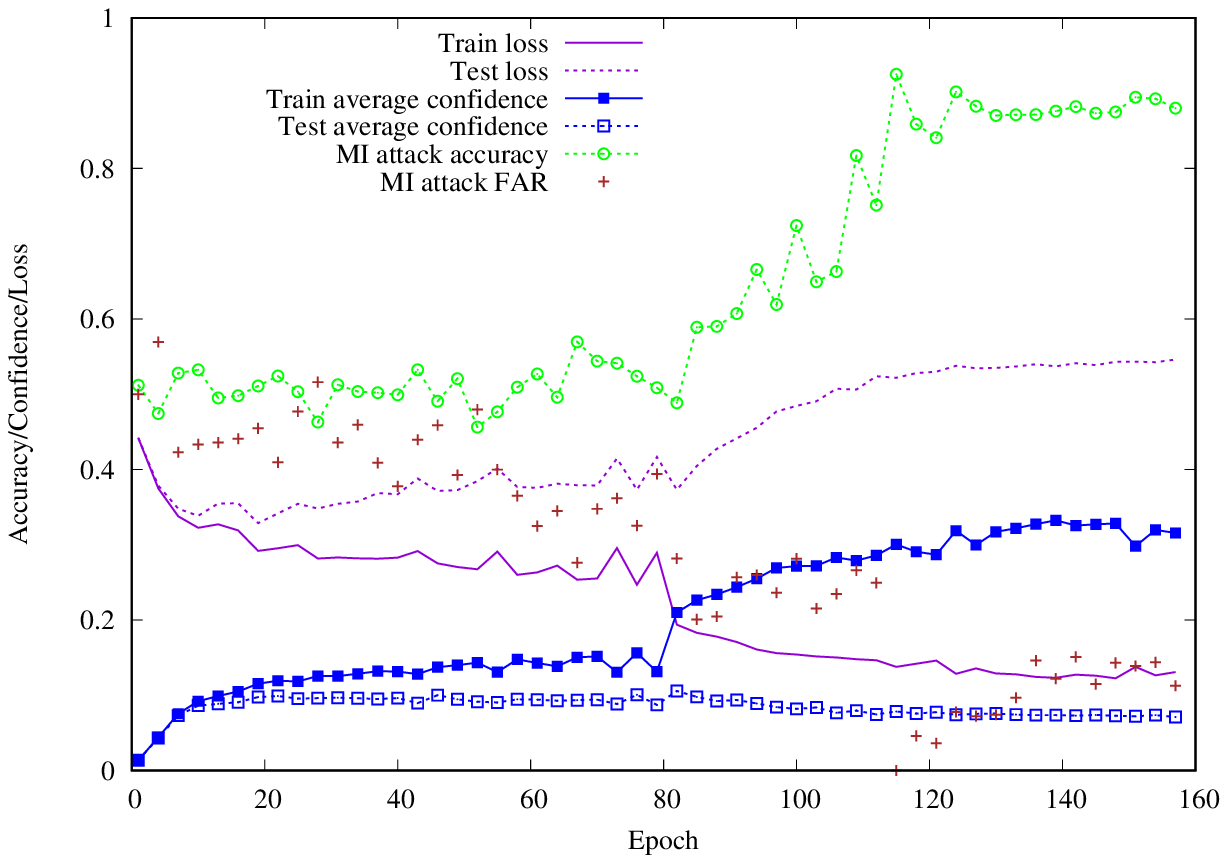}}\\
\end{tabular}
\end{tabularx}
\caption{Training progress and MI attack on CIFAR-100 for ResNet model}
\label{cifar100-train-progress-resnet}
\end{figure*}

\begin{figure*}[h]
\def\tabularxcolumn#1{m{#1}}
\begin{tabularx}{\linewidth}{@{}cXX@{}}
\begin{tabular}{ccc}
\subfloat[All dataset]{\includegraphics[width=0.3\linewidth]{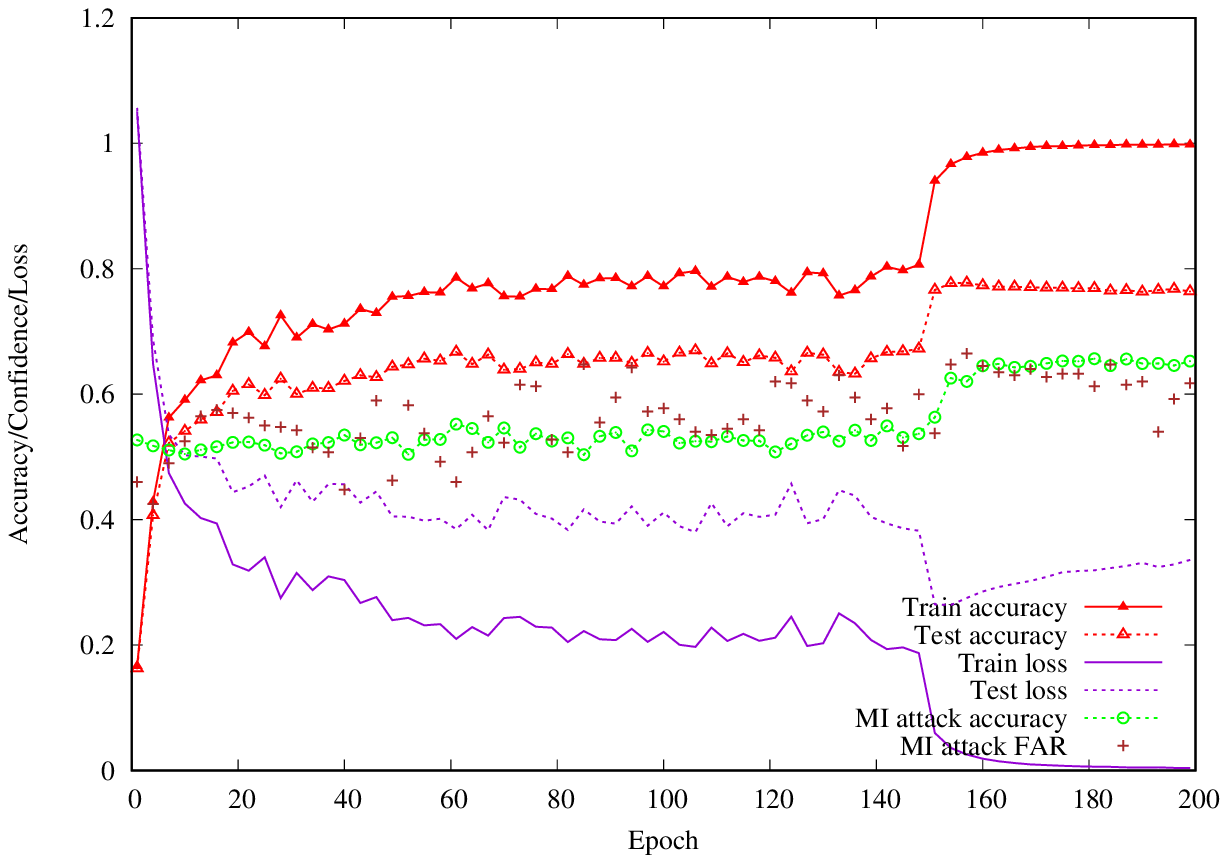}} 
& \subfloat[Correctly classified samples]{\includegraphics[width=0.3\linewidth]{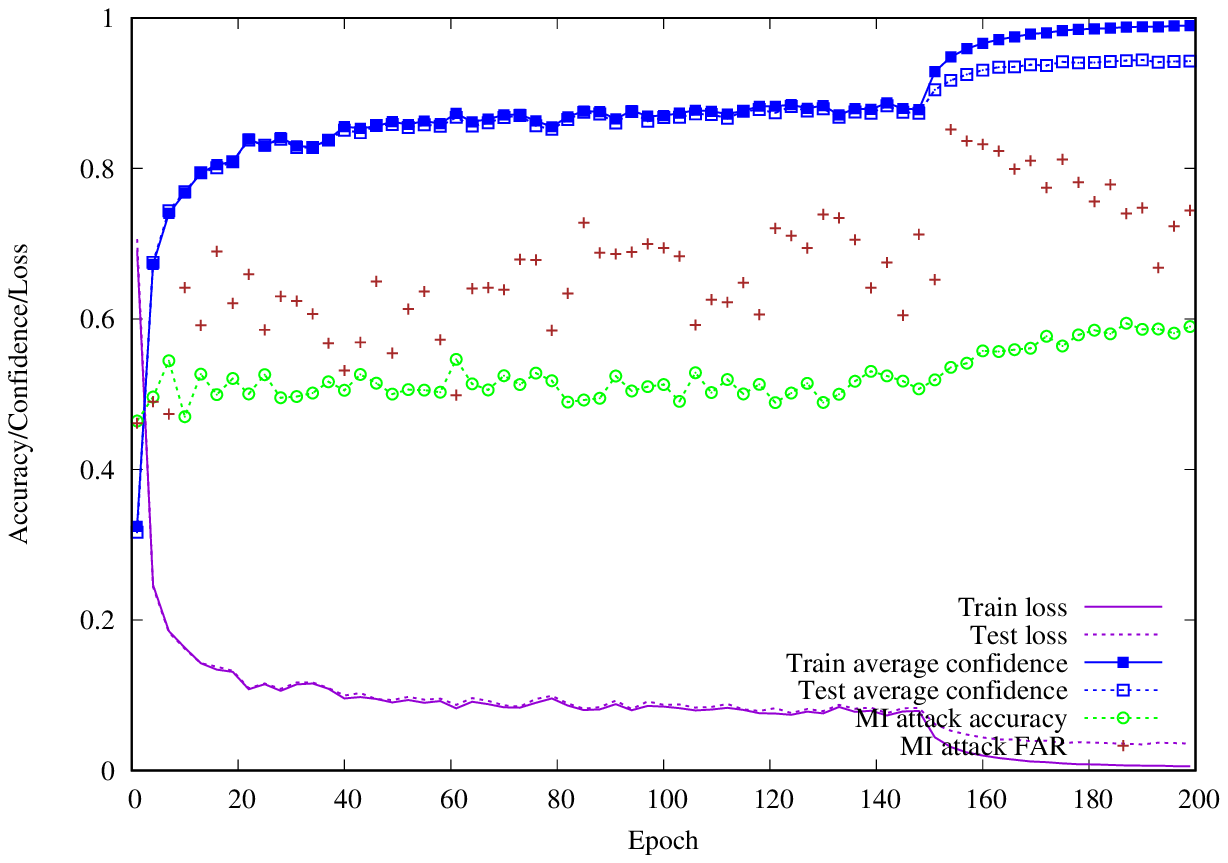}}
& \subfloat[Mis-classified samples]{\includegraphics[width=0.3\linewidth]{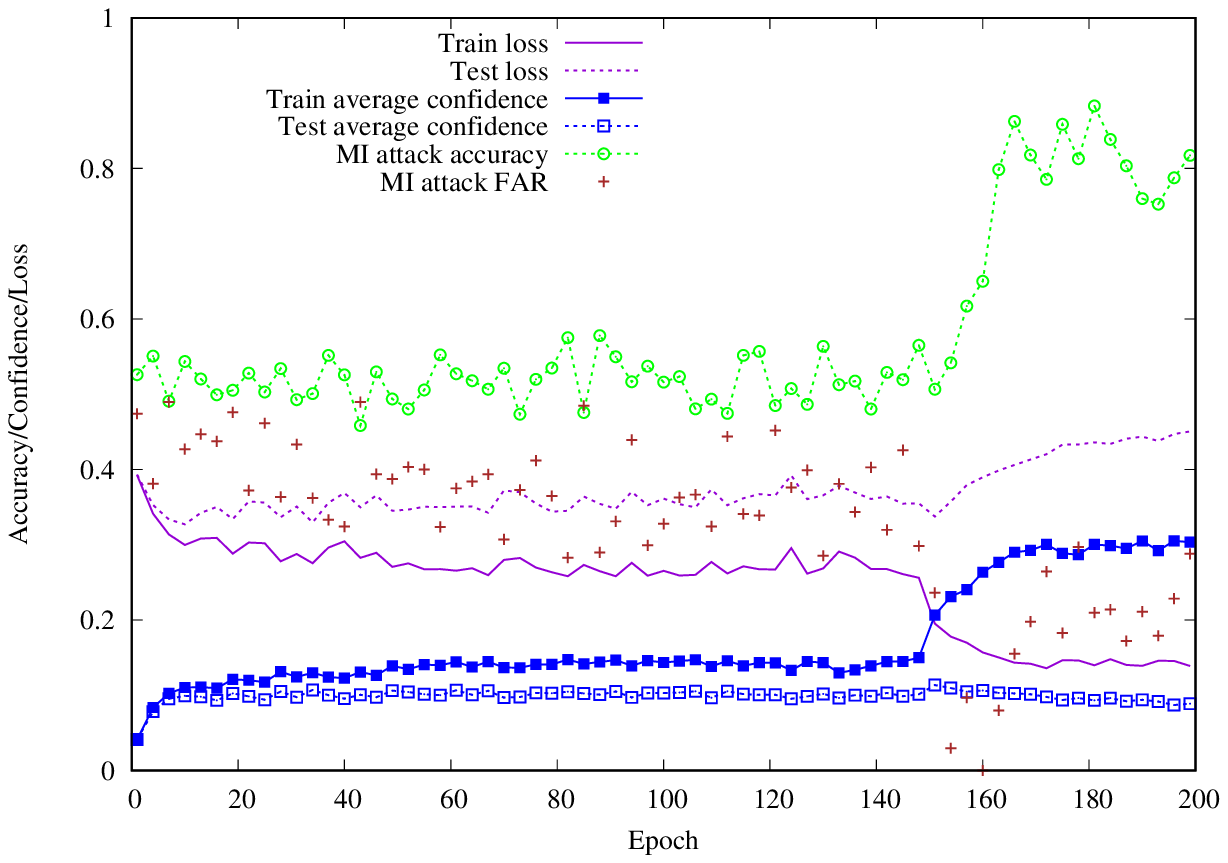}}\\
\end{tabular}
\end{tabularx}
\caption{Training progress and MI attack on CIFAR-100 for DenseNet model}
\label{cifar100-train-progress-densenet}
\end{figure*}

In this section, we analyze the effect of overfitting on membership inference. Note that extremely overfitted models have no practical use in reality. The goal of this section is to show that the overfitted models may behave differently than well-trained models. As a result, researchers should avoid using overfitted models for MI attack and generalize them to well-trained practical models. To show the effect of overfitting, we train AlexNet, ResNet, and DenseNet models for a fixed amount of epochs on CIFAR-10 and CIFAR-100. We use the same training parameters as used by Wei Yang\footnote{https://github.com/bearpaw/pytorch-classification}. We launch MI attack based on confidence values on various epochs during the training.
The results are shown in Figure \ref{cifar10-train-progress-alexnet}, \ref{cifar10-train-progress-resnet}, \ref{cifar10-train-progress-densenet}, \ref{cifar100-train-progress-alexnet}, \ref{cifar100-train-progress-resnet}, and \ref{cifar100-train-progress-densenet}.

As shown in Figure \ref{cifar10-train-progress-alexnet}(a), the model starts overfitting around epoch 80, when the loss function for the test set stops improving. It is clear that all MI attacks before the epoch 80 suffers from low accuracy (almost similar to random guess) and high FAR, on both correctly classified samples (Figure \ref{cifar10-train-progress-alexnet}(b)) and misclassified samples (Figure \ref{cifar10-train-progress-alexnet}(c)). On the other hand, as the target model start overfitting, the performance of MI attacks increases over misclassified samples (Figure \ref{cifar10-train-progress-alexnet}(c)). This phenomenon is more evident on other models, such as ResNet (Figure \ref{cifar10-train-progress-resnet}(c)). However, overfitting does not significantly improve MI attacks on correctly classified samples. Note than one should consider the number of misclassified training (member) samples to evaluate if the high performance MI attacks on misclassified samples have any real impact. The reason is that as target models overfit, the number of misclassified training samples approaches zero. In most cases, after epoch 160, there are only a handful of misclassified training samples. In other words, even a successful MI attack on an overfitted model only reveals the membership status of a handful of training samples. In any case, adopting a simple technique, such as early stopping, can even eliminate such as possibility.

\end{document}